\documentclass[preprint,12pt]{elsarticle}

\usepackage{amsmath,amssymb,amsfonts}
\usepackage{gensymb}
\usepackage{algorithmic}
\usepackage{graphicx}
\usepackage{textcomp}
\usepackage[table]{xcolor}
\usepackage{algorithm}
\usepackage{url}
\usepackage{nicematrix}
\usepackage{multirow}
\usepackage{subcaption}
\usepackage{float}
\usepackage{flushend}

\usepackage{rotating}
\usepackage{tabularx}
\usepackage{ragged2e}
\newcolumntype{L}{>{\RaggedRight\arraybackslash}X}

\usepackage{tikz}
\usetikzlibrary{arrows.meta,angles,quotes,calc}

\journal{Robotics and Automation Systems}

\begin{document}

\begin{frontmatter}

  \title{Dynamic UAV-based search operations using \\ probabilistic diffusion modeling of \\ Man Overboard incident victims} %

  \author{Dimosthenis Angelis} %

  \affiliation{organization={Technical University of Denmark},%
    city={Kgs. Lyngby, Denmark},
    postcode={2800},
    country={Denmark}}

  \author{Evangelos Boukas} %

  \affiliation{organization={Technical University of Denmark},%
    city={Kgs. Lyngby, Denmark},
    postcode={2800},
    country={Denmark}}

  \begin{abstract}
    {More than $70\%$ of the people that fell overboard cruise ships in the period 2010-2019 lost their lives \cite{stat4}.} This paper presents a strategy for reliably predicting the area a person may be in after a man overboard incident, and describes in detail the search methods to find them utilizing UAV technology. The search area prediction method employs an Extended Kalman Filter that capitalizes on the information from the Leeway model to track the missing person in the sea by taking into account the uncertainty of the movement of the person and the weather conditions in the area. Then, a UAV uses this information to search for the person. Five different methods for searching in this dynamic area are presented and evaluated - the Zigzag, the Boustrophedon, the Spiral, the Probability Informed Search and the Improved Probability Informed Search (IPIS) methods. {The IPIS method provides success rate of over 80\% on average for finding a person, even if the UAV initiates the search mission 20 minutes after the man overboard incident and even assuming a detection method with a success rate of 30\%. }All code for the simulation environment and the evaluation of the methods is available on our GitHub page at \url{https://github.com/diangeli/pdms-man-overboard}.
  \end{abstract}

  \begin{highlights}
    \item We implemented a novel position estimator for MOB incidents in open sea that can accurately estimate a search area where a missing person is in, and can run onboard an autonomous vehicle.
    \item We implemented 5 different search methods for dynamically expanding areas, 3 previously proposed naive methods and 2 newly formulated probability informed methods, and evaluated them in the context of fast emergency response for MOB incidents.
    \item We created a methodology to increase the success rate of finding a missing person in water utilizing a UAV to more than {$80\%$}, even if the UAV starts a search mission $20$ minutes after the MOB incident.
  \end{highlights}

  \begin{keyword}
    Maritime Search and Rescue \sep
    Unmanned Aerial Vehicles \sep
    Dynamic Area Path Planning \sep
    Probabilistic Modeling
  \end{keyword}

\end{frontmatter}

\section{Introduction}

Accidents such as people falling overboard vessels, also known as man-overboard (MOB) incidents, can happen during transport and leisure activities at sea, endangering the lives of the people involved. A study that was conducted in 2020 \cite{stat0} reported 623 fatalities, out of which 23\% were accounted to MOB incidents. The study reported that MOB incidents were the most common reason for fatalities at sea. Most of these fatalities were passenger related. In addition, the "Annual overview of marine casualties and incidents 2024" \cite{stat2} reported 26596 marine accidents in the period 2014 to 2023, leading to 7604 human injuries and 650 work-related fatalities. The Cruise Lines International Association (CLIA), reported that only 28.2\% of the people that fell overboard vessels in the period 2009-2019, survived \cite{stat4}. This shows limitations of the search and rescue procedures that are currently employed in the maritime industry.

\begin{figure}[h]
  \centering
  \includegraphics[width=100mm]{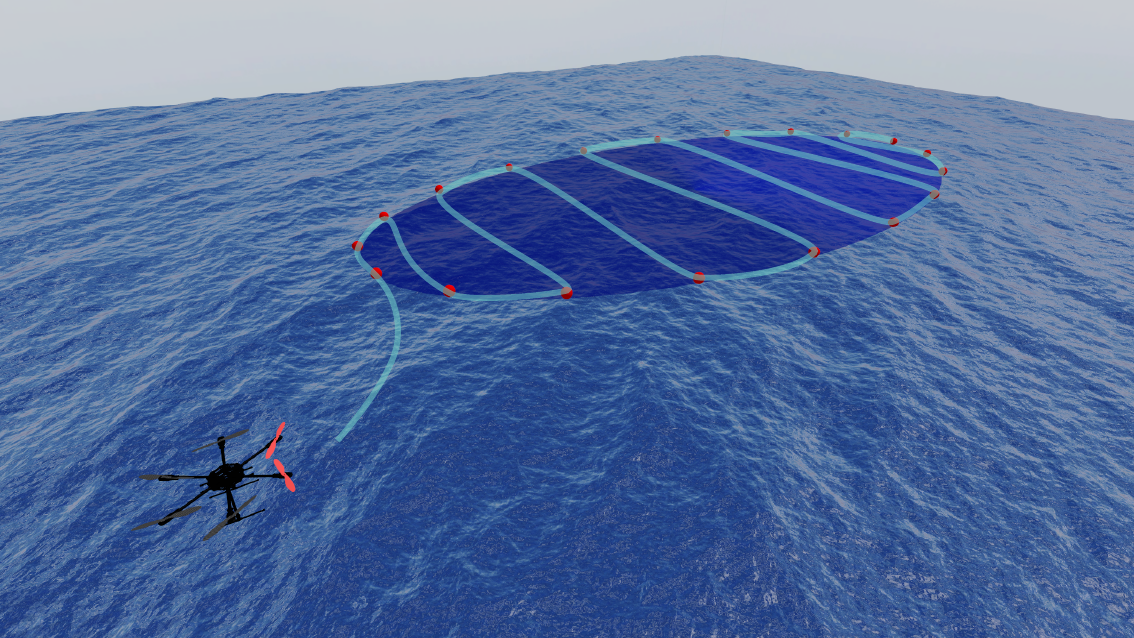}
  \caption{Simulation of UAV searching for people at sea in Gazebo. The blue ellipse represents the predicted search area, the red spots are the calculated nodes that the UAV has to traverse through, and the turquoise line is the calculated path that the UAV has to follow to completely cover the area, using in this example the boustrophedon search method.}
  \label{fig:intro-drawing}
\end{figure}

We tackle this problem by enhancing the search procedure of MOB incidents. Our method aims to solve the problem of de facto late emergency response to an MOB incident, by deploying an Unmanned Aerial Vehicle (UAV) onboard the vessel. We have developed an Extended Kalman Filter (EKF) based method to predict the location of people who have fallen overboard ships. The EKF uses the leeway information of a person drifting in water, and incorporates the wind and current sensor measurements provided by the vessel, to accurately predict the area they might be in. An expanding uncertainty ellipse (${\ge}{3\sigma}$) is produced which is used as a search area. This ellipse is updated every second, making the search area dynamically expand through time. We then search this area using five different search methods. These methods are designed in a way that makes it possible to search for the person, while also following the ellipse. An illustration of the procedure can be seen in Figure \ref{fig:intro-drawing}.

Our approach includes key features that make it well-suited for search and rescue missions with a UAV. The EKF is able to model the movement of a person in the water accurately using the Leeway model. The uncertainty of the wind and current measurements are incorporated in the estimator to increase its performance. The method minimizes the computation power needed to estimate the probable area the missing person might be in, which allows it to run onboard the UAV. The search methods are able to effectively search for people in the predicted search areas. On one hand, the naive methods complete multiple passes for the search area, resulting in high success rates at the cost of longer search times. On the other hand, the probability-informed methods utilize the probability maps that are produced from the estimator to increase the success rate in the early stages of the search.

The main contributions of the paper are:
\begin{itemize}
  \item a novel position estimator for MOB incidents in open sea that can accurately estimate a search area where a missing person is in, and can run onboard an autonomous vehicle,
  \item proposal of a probability informed search method in the context of MOB incidents,
  \item adaptation of three previously proposed search methods for static search areas to dynamic search areas.
\end{itemize}

\section{Related work}
\label{related-work}

\subsection{Drift of person in water}

Identifying the drift of a person in water is essential to the success of a SAR mission. The Leeway model is used to determine the position of a passively drifting object in water. It was proposed by Allen and Plourde in 1999 \cite{leeway1}, and it statistically identifies the wind induced drift of free-floating bodies in water. The method was later refined by including the force of wind acting on the drifting object surface \cite{leeway2}. The characteristic parameters for different kind of objects were experimentally calculated based on field-testing \cite{leeway3}. Leeway simulations have previously been used for tracking oil spills, lifeboats, human bodies, and marine litter under wind conditions \cite{leeway3, leewaysim1,leewaysim2,leewaysim5,leewaysim6,leewaysim7,leeway4}. We define the search area as a weather-based dynamic ellipse, which expands over time. The expansion of the search area reduces the probability of finding the person of interest. Both our ground truth method and estimation method capitalize on the Leeway model to define and continuously update the search area.

Previous studies have built upon the properties of the Leeway model, resulting in stochastic models that define search areas \cite{leeway4, RWPT4, leeway7, RL1}. These models use deterministic wind and current conditions and generate potential fields to estimate the search area. The methods are validated based on previous data and experiments at sea. In our approach, we define the wind and current information as Gaussian distributions, based on the data provided by the vessel at the time of the MOB incident. This eliminates the need {for} external measurements from equipment that is not available to the vessel.

OpenDrift is a framework used for trajectory estimation of diverse objects in the ocean and the atmosphere \cite{opendrift}. It is a general purpose Lagrangian particle modeling framework for fast development and deployment of particle methods that use scalar or vector forcing fields. OpenDrift performance was experimentally validated in \cite{leeway8}. In \cite{RWPT4}, OpenDrift was used to determine the drifting of people in the water using the empirical data collected in \cite{leeway1}. Our simulation environment uses a Monte Carlo random particle tracking model, based on the same principles, for continuously estimating the movement of people in the area. This method has been previously used \cite{RWPT4,RWPT1,RWPT2,RWPT3} to conduct SAR missions during maritime emergencies. The positions of the people are used as ground truth data to tune and validate our estimation method.

\begin{sidewaystable}[p]
  \centering
  \begin{tabularx}{\textwidth}{%
      p{3cm}
      p{5cm}
      p{4cm}
      p{4cm}
      p{1cm}     
    }
    \hline
    \textbf{Name}
      & \textbf{Description}
      & \textbf{Method}
      & \textbf{Objects tracked}
      & \textbf{Online} \\
    \hline
    GNOME \cite{gnome}
      & Eulerian/Lagrangian oil-spill model
      & Lagrangian elements within continuous flow fields
      & Oil spills
      & No \\
    MEDSLIK-II \cite{medslik1, medslik2}
      & Lagrangian oil-spill model for short-term forecasting
      & Particle advection, diffusion and weathering processes
      & Oil spills
      & No \\
    TRACMASS \cite{tracmass}
      & Mass-conserving Lagrangian trajectory code
      & Analytical integration through grid cells
      & Passive particles in water
      & No \\
    Parcels \cite{parcels}
      & Python Lagrangian analysis library
      & Lagrangian ocean analysis
      & Passive particles in water
      & No \\
    ARIANE \cite{ariane1, ariane2}
      & 3D streamline diagnostic tool
      & Analytical C-grid advection
      & Water parcel streamlines
      & No \\
    CMS \cite{cms}
      & Probabilistic, multiscale model for diffusion processes in the ocean
      & Combination of stochastic Individual-Based Models and Lagrangian frameworks
      & Larvae, pollutants, plastics
      & No \\
    OpenDrift \cite{opendrift}
      & Modular Python framework for generic drift modeling
      & Monte Carlo Lagrangian particle ensembles
      & Oil, debris, larvae, SAR objects
      & No \\
    SAROPS \cite{sarops}
      & Operational SAR planning system
      & Ensemble stochastic trajectories with leeway
      & SAR targets (rafts, persons)
      & Yes \\
    \hline
  \end{tabularx}
  \caption{Drift‐prediction systems for objects in the water}
  \label{tab:prediction-methods}
\end{sidewaystable}

Table \ref{tab:prediction-methods} summarizes the systems currently used to predict the drift of objects in water. Most of the referenced systems are designed for offline simulations and predictions of incidents. They have been validated on real incidents and are currently being used for predictions. Search and Rescue Optimal Planning System (SAROPS) \cite{sarops} is the only system in Table \ref{tab:prediction-methods} specifically designed to run online by fetching real time weather data. It also employs the Leeway model, and uses a particle filter to predict time-evolving probability density maps. In our solution, we use an Extended-Kalman-Filter-based approach to estimate a time-evolving probability search area, making it possible to run the estimation method onboard the UAV during the search using minimal computational resources.

\subsection{Search methods for open maritime areas}
Different methods for full coverage of static search areas have been proposed for search and rescue missions in maritime environments \cite{search0}. The sector search path was proposed for small search areas using surface vehicles. After selecting a central node, the search vehicle executes the search mission which involves straight paths of diameter $R$ followed by right turning circular arcs with a radius R and a central angle of 120 degrees. The search mission is completed when the whole search area is explored. The parallel search path, otherwise known as lawn-mower or boustrophedon method \cite{search1}, was proposed to search in larger and more uncertain areas, when the sea currents are not known. The search vehicle follows parallel paths with width distance smaller than $2R_d$, where $R_d$ is the effective area of the search system. The extended square search path, also known as spiral search path, was proposed for searching of people when the current direction and escape direction of the victim is unknown. The search vehicle starts from the last known node and navigates in a gradually expanding outward manner. The width of the search route is also $2R_d$, as in the parallel search path method. We formally formulate these methods in the context of aerial vehicle search for dynamically expanding search areas and compare them with our proposed methods.

In \cite{feraru}, an overall search procedure is outlined for an MOB incident using a UAV. The search area is defined by merging the probability ellipses over a period of time - starting from the time of arrival of the search vehicle to the search area and finishing when the area is fully explored, or the trip is not viable due to power constraints. The boustrophedon method was used to search for the missing person in the search area. In this paper, we optimize the search by only choosing nodes inside the dynamic probability ellipse at each time step.

In \cite{dimos}, a UAV aimed for fully autonomous search and rescue missions in maritime environments was presented. The UAV is to be stationed on the vessel, reducing the response time after an MOB incident occurs. It is designed to sustain an average speed of 20 meters per second, while withstanding the harsh weather conditions of the sea. Its sensor suite comprises an RGB, a near infrared and a thermal camera to detect people in the sea. We use the characteristics of this UAV to evaluate the developed search methods in our simulation experiments.

{In \cite{RL1}, a framework for maritime SAR coverage path planning for surface vehicles is introduced. The framework employs a drift trajectory module that predicts a hierarchical probability environment map based on the Leeway model and Monte Carlo particle simulation. The maps are used by a reinforcement learning model that incorporates the probability of containment and the search history of the vehicle, to optimize the path coverage of the predicted areas. Similarly, our prediction method is using the same underlying model to predict efficiently the areas that may contain people, but our proposed probability informed methods offer a deterministic and robust alternative to the proposed learning method for UAVs.}

{Using multiple vehicles for maritime SAR missions is usually necessary due to the dynamic nature of the sea and how fast the probable area of containing a person can become. In \cite{RL3}, a resource allocation decision model is proposed to determine the optimal resource allocation scheme for a maritime SAR scenario. The proposed method combines a genetic algorithm with reinforcement learning to address the complexity of marine environments. In \cite{RL2}, a hybrid optimization framework is proposed for the coordination of multiple UAVs across several rescue centers, in the context of large-scale maritime SAR operations. The problem is formulated as a mixed integer programming model that minimizes the total search time under battery and routing feasibility constraints. To solve this problem, Q-learning is integrated in a genetic algorithm to adaptively manage the dynamic nature of the mission. Our proposed solution focuses on the early response to the MOB incident during the first minutes, eliminating the need for extensive multi-agent SAR operations.}

\section{MOB estimation}
\label{sec:mob-estimation}
The motion of an object at sea can be determined by the forces that act on it. These forces are generated by the winds, currents and waves in the vicinity of the object. The speed of a free-floating object can be defined as:

\begin{equation}
  V = V_{C} + V_{L} + V_{W},
  \label{eq:forces}
\end{equation}
where $V_{C}$ is the current induced drift, $V_{L}$ is the wind induced drift, and $V_{W}$ is the wave induced drift. The current induced drift speed is equal to the current speed at the surface level of the sea\cite{leeway6}. In the case of a person in water, the wave induced drift is disregarded due to the small size of humans \cite{leeway4}.

\subsection{The Leeway model}
\label{sec:leeway-model}
The Leeway model utilizes the wind speed and direction at a reference node of 10 meters above the sea level, to estimate the wind induced drifting of a free-floating body in water. Allen and Plourde \cite{leeway1} observed an almost linear relationship between wind speed and downwind and crosswind leeway components, thus they decoupled the downwind and crosswind components of the drift, and formulated them as:

\begin{equation}
  L_d = a_d W_{10} + b_d + \epsilon_d,
  \label{eq:downwind_leeway}
\end{equation}

\begin{equation}
  L_{c}^{+} = a_{c}^{+} W_{10} + b_{c}^{+} + \epsilon_c^+,
  \label{eq:crosswind_leeway1}
\end{equation}

\begin{equation}
  L_{c}^{-} = a_{c}^{-} W_{10} + b_{c}^{-} + \epsilon_c^-,
  \label{eq:crosswind_leeway2}
\end{equation}
{where $L_{d}, L_{c}$ are the downwind and crosswind leeway components, $W_{10}$ is the wind speed at the $10$ meter reference node, $a_{d}, a_{c}$ are the downwind and crosswind slopes, $b_{d}, b_{c}$ are the downwind and crosswind standard deviations for the unconstrained linear regression, and $\epsilon_{d}, \epsilon_{c}$ are the regression errors. To describe the uncertainty in field experiments, the slopes and offsets were adjusted by adding a perturbation term \cite{leeway3, leeway9}:}

{
  \begin{equation}
    L_d = (a_d + \frac{\epsilon_d}{20}) W_{10} + (b_d + \frac{\epsilon_d}{2}),
  \label{eq:downwind_leeway_with_perturbation}
  \end{equation}
}

{where $\epsilon_d = S_{yx} \cdot \text{norm}$, norm is a random number that follows a normal distribution $\mathcal{N} (0, 1)$, and $S_{yx}$ is the standard error term of the linear regression of the leeway components versus 10m wind speed.}

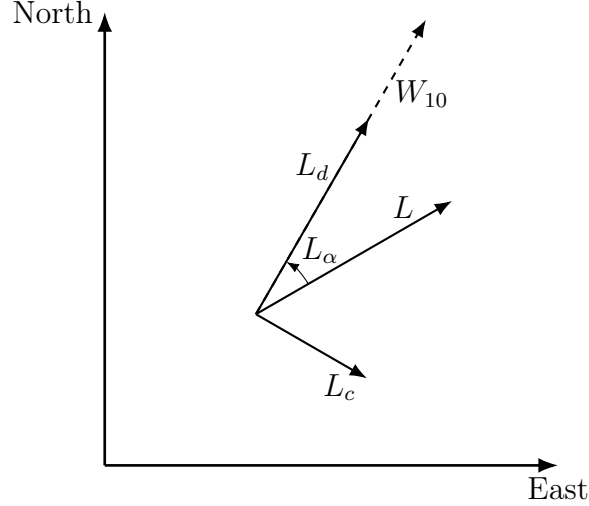
\begin{figure}[h]
  \centering
  \begin{tikzpicture}[>=Latex]
    \draw[->,line width=1pt] (-2,-2) -- (-2,4) node[left] {North};
    \draw[->,line width=1pt] (-2,-2) -- (4,-2) node[below] {East};

    \coordinate (O)  at (0,0);
    \coordinate (Dd) at ({3*cos(60)},{3*sin(60)});
    \coordinate (Wd) at ({4.5*cos(60)},{4.5*sin(60)});
    \coordinate (L)  at ({3*cos(30)},{3*sin(30)});
    \coordinate (Cc) at ({1.7*cos(-30)},{1.7*sin(-30)});

    \draw[->,thick]        (O) -- (Dd)  node[near end,left]  {$L_d$};
    \draw[->,dashed,thick] (O) -- (Wd)  node[near end,right] {$W_{10}$};
    \draw[->,thick]        (O) -- (L)   node[near end,above] {$L$};
    \draw[->,thick]        (O) -- (Cc)  node[near end,below] {$L_c$};

    \pic[draw, ->, angle radius=8mm, angle eccentricity=1.5, "$L_\alpha$"] {angle = L--O--Dd};

  \end{tikzpicture}
  \caption{The leeway ($L$) consists of the downwind ($L_d$) and crosswind ($L_c$) components.}
  \label{fig:leeway}
\end{figure}

The downwind and crosswind components of the leeway can be seen in Figure \ref{fig:leeway}. The total motion of an object is then defined as:

\begin{equation}
  x_{t} - x_{0} = \int_{0}^{T} V(t) \, dt = \int_{0}^{T} \left( V_{C}(t) + V_{L}(t) \right) dt
\end{equation}

For a person in water, the downwind slope is $a_d = 1.93\%$ with standard deviation of $0.083 m s^{-1}$, the positive crosswind slope is $a_{c}^{+} = 0.51\%$ with standard deviation of $b_{c}^{+} = 0.067 m s^{-1}$, and the negative crosswind slope is $a_{c}^{-} = 0.51\%$ with standard deviation of $b_{c}^{-} = 0.067 m s^{-1}$ \cite{RWPT4}.

A free-floating body in water can move with either a positive or a negative crosswind component. This is affected by the shape and orientation of the object. The change between the positive and negative crosswind is called jibing and is critical to the estimation of the position of an object in water. In the case of a person, the jibing frequency is $50\%$ \cite{leeway2}.

\subsection{Monte Carlo random particle method}
\label{montarlo}
Lagrangian or particle tracking models have previously been used for general purpose tracking of objects at sea. A number of particles are initialized at a specific location, and are subjected to current and wind effects on each time step. The particles are considered passive, meaning that the people move solely on the external forces that are applied to them. For each time step, a distribution of the particles can be calculated, which defines the search area of the mission for that time step.

In this work, we use a Monte Carlo random particle model \cite{montecarlo1} to predict the drift trajectory of multiple people for Lagrange particle tracking. A number of particles $N_p$ are initialized at a position $x_0$. For each time step, the motion of each particle is calculated by:

\begin{equation}
  \mathbf{x}_t = \mathbf{x}_{t-1} + \mathbf{v} dt,
  \label{eq:speed_timestep}
\end{equation}
where:
{\begin{equation*}
  \mathbf{v} = \begin{bmatrix}
    V_C\sin{\theta_c} + L_d\sin{\theta_w} - L_c\cos{\left(\theta_w \pm \pi/2\right)} \\
    V_C\cos{\theta_c} + L_d\cos{\theta_w} + L_c\sin{\left(\theta_w \pm \pi/2\right)}
  \end{bmatrix}.
\end{equation*}}

Variables $\theta_c$ and $\theta_w$ are the current direction and wind direction, respectively. The parameters for the leeway equations are sampled based on the values that are described in chapter \ref{sec:leeway-model} for each individual particle. The sign for the crosswind leeway component is determined randomly for each particle on each time step. This happens because the parameters for the positive and negative crosswind components are symmetrical and the jibing frequency for the person in water is $50\%$.

The wind speed is defined as $V_w \sim \mathcal{N}(\mu_{V_w}, \sigma^2_{V_w})$ with direction $\theta_w \sim \mathcal{N}(\mu_{\theta_w}, \sigma^2_{\theta_w})$, and the current speed as $V_C \sim \mathcal{N}(\mu_{V_c}, \sigma^2_{V_c})$ with direction $\theta_c \sim \mathcal{N}(\mu_{\theta_c}, \sigma^2_{\theta_c})$. By propagating these distributions through equation \ref{eq:downwind_leeway_with_perturbation}, we calculate the downwind and crosswind leeway distributions:
{
  \begin{equation}
    L_d \sim \mathcal{N}\!\left(\mu_{L_d}, \sigma^2_{L_d}\right)
    = \mathcal{N}\!\left(
    a_d \mu_{V_w} + b_d,\;
    a_d^2 \sigma^2_{V_w}
    + \frac{S_{yx}}{400}
    \left[(10 + \mu_{V_w})^2 + \sigma^2_{V_w}\right]
    \right)
  \end{equation}
  \begin{equation}
    L_c \sim \mathcal{N}\!\left(\mu_{L_c}, \sigma^2_{L_c}\right)
    = \mathcal{N}\!\left(
    a_c \mu_{V_w} + b_c,\;
    a_c^2 \sigma^2_{V_w}
    + \frac{S_{yx}}{400}
    \left[(10 + \mu_{V_w})^2 + \sigma^2_{V_w}\right]
    \right)
  \end{equation}
}

The values for the mean and variance of the leeway distributions are defined by the data that are collected by the ship. Since the proposed method is aimed to solve the fast response between the MOB incident and the search mission, we assume that the weather data the ship collects are highly correlated with the data from the area. On each time step, the wind speed, wind direction, current speed and current direction are sampled from their respective distributions.

The Monte Carlo random particle model is used to produce ground truth data for the drifting positions of people in water. The use of many particles results in a very good estimation of the search area on each time step, but gives rise to a very computationally expensive system. A performance comparison of the Monte Carlo random particle model, for different number of particles, and the Extended Kalman Filter method is shown in Table \ref{tab:performance-comparison}.

\subsection{Extended Kalman Filter-based tracking of man-overboard person}
\label{ekf}
In this section, we introduce an Extended Kalman Filter based approach to estimate the position of people in water. We define the problem as a tracking of a person in water with a state vector as:
\begin{equation}
  \mathbf{x}_{k} = \begin{bmatrix}
    \mu_{x}        \\
    \mu_{y}        \\
    \mu_{V_c}      \\
    \mu_{L_d}      \\
    \mu_{L_c}      \\
    \mu_{\theta_c} \\
    \mu_{\theta_w}
  \end{bmatrix},
\end{equation}
where $\mu_{x}, \mu_{y}$ are coordinates of the mean position of the people in water. We initialize the covariance matrix $\mathbf{P}_k$ using the information collected from the ship as:

\begin{equation}
  \mathbf{P}_k = \mathbf{I}_{7x7} \begin{bmatrix}
    0.1                 \\
    0.1                 \\
    \sigma_{V_c}^2      \\
    \sigma_{L_d}^2      \\
    \sigma_{L_c}^2      \\
    \sigma_{\theta_c}^2 \\
    \sigma_{\theta_w}^2
  \end{bmatrix}^T.
\end{equation}

In the regular EKF approach, equation \ref{eq:speed_timestep} is used to produce the state transition matrix. Since $V_c, V_w, \theta_c, \theta_w$ are defined as Gaussian distributions, equation \ref{eq:speed_timestep} cannot be used as is for the prediction part of the estimation reliably. {
  To tackle this issue, we calculate the analytical mean value of the speed of a person in the water. For the positive crosswind component we calculate the x and y velocity component as:  
  \begin{align}
    V_x^{+} &= V_c \sin\theta_c + L_d \sin\theta_w - L_c \cos\theta_w, \label{eq:vx_plus_def} \\
    V_y^{+} &= V_c \cos\theta_c + L_d \cos\theta_w + L_c \sin\theta_w, \label{eq:vy_plus_def}
  \end{align}
  By applying the expectation operator and factorization we get: 
  \begin{align}
    \mu_{V_x^{+}}
    &= \mathbb{E}[V_x^{+}] \notag \\
    &= \mathbb{E}[V_c\sin\theta_c]
    + \mathbb{E}[L_d\sin\theta_w]
    - \mathbb{E}[L_c\cos\theta_w] \notag \\
    &=\mathbb{E}[V_c]\mathbb{E}[\sin\theta_c]
    + \mathbb{E}[L_d]\mathbb{E}[\sin\theta_w]
    - \mathbb{E}[L_c]\mathbb{E}[\cos\theta_w],
    \label{eq:vx_plus_expectation} \\
    \mu_{V_y^{+}}
    &= \mathbb{E}[V_y^{+}] \notag \\
    &= \mathbb{E}[V_c\cos\theta_c]
    + \mathbb{E}[L_d\cos\theta_w]
    + \mathbb{E}[L_c\sin\theta_w] \notag \\
    &= \mathbb{E}[V_c]\mathbb{E}[\cos\theta_c]
    + \mathbb{E}[L_d]\mathbb{E}[\cos\theta_w]
    + \mathbb{E}[L_c]\mathbb{E}[\sin\theta_w].
    \label{eq:vy_plus_expectation}
  \end{align}
  Since $\theta_w$ and $\theta_c$ are normally distributed random variables, we can calculate the expectation of their sine and cosine functions as: 
  \begin{align}
    \mathbb{E}[\cos\theta] &= e^{-\frac{\sigma_{{\theta_c}}^2}{2}}\cos\mu_\theta, \label{eq:cos_expectation} \\
    \mathbb{E}[\sin\theta] &= e^{-\frac{\sigma_{{\theta_c}}^2}{2}}\sin\mu_\theta. \label{eq:sin_expectation}
  \end{align}
  By applying equations \ref{eq:cos_expectation} and \ref{eq:sin_expectation} to \ref{eq:vx_plus_def} we get mean velocity: 
  \begin{align}
    \mu_{V_x^{+}} =
    e^{-\frac{\sigma_{{\theta_c}}^2}{2}} \, \mu_{V_c} \sin\mu_{\theta_c}
    + e^{-\frac{\sigma_{{\theta_w}}^2}{2}}
    \left(
    \mu_{V_{ld}} \sin\mu_{\theta_w}
    - \mu_{V_{lc}} \cos\mu_{\theta_w}
    \right),
    \label{eq:vx_plus_code} \\
    \mu_{V_y^{+}} =
    e^{-\frac{\sigma_{{\theta_c}}^2}{2}} \, \mu_{V_c} \cos\mu_{\theta_c}
    + e^{-\frac{\sigma_{{\theta_w}}^2}{2}}
    \left(
    \mu_{V_{ld}} \cos\mu_{\theta_w}
    + \mu_{V_{lc}} \sin\mu_{\theta_w}
    \right).
    \label{eq:vy_plus_code}
  \end{align}
  Similarly, we can get the negative crosswind leeway components as: 
  \begin{align}
  \mu_{V_x^{-}} &=
    e^{-\frac{\sigma_{{\theta_c}}^2}{2}} \, \mu_{V_c} \sin\mu_{\theta_c}
    + e^{-\frac{\sigma_{{\theta_w}}^2}{2}}
    \left(
    \mu_{V_{lc}} \cos\mu_{\theta_w}
    + \mu_{V_{ld}} \sin\mu_{\theta_w}
    \right),
    \label{eq:vx_minus_code} \\[6pt]
    \mu_{V_y^{-}} &=
    e^{-\frac{\sigma_{{\theta_c}}^2}{2}} \, \mu_{V_c} \cos\mu_{\theta_c}
    + e^{-\frac{\sigma_{{\theta_w}}^2}{2}}
    \left(
    \mu_{V_{ld}} \cos\mu_{\theta_w}
    - \mu_{V_{lc}} \sin\mu_{\theta_w}
    \right).
    \label{eq:vy_minus_code}
  \end{align}
}

Equations \ref{eq:vx_plus_code}-\ref{eq:vy_minus_code} are used to compute the state transition matrix on each time step. The equations are used alternating on each time step, so that the filter can capture the jibing frequency of the person in water. The state transition matrix is calculated on each time step as:
\begin{equation}
  f(\hat{\mathbf{x}}_{k-1}) = \begin{bmatrix}
    \hat{x}_{k-1}+\mu_{V_x} dt \\
    \hat{y}_{k-1}+\mu_{V_y} dt \\
    \mu_{V_c}                  \\
    \mu_{L_d}                  \\
    \mu_{L_c}                  \\
    \mu_{\theta_c}             \\
    \mu_{\theta_w}
  \end{bmatrix}
\end{equation}

The standard deviation of the drifting speed of a person in water cannot be calculated analytically, due to the Gaussian terms that pass through the sine and cosine functions. {We use equations \ref{eq:vx_plus_code}-\ref{eq:vy_minus_code} for the calculation of the Jacobian matrices for the positive and negative crosswind components as:}

{
  \begin{equation}
    \mathbf{F}_k^{+} =
    \makebox[\linewidth][c]{%
    \resizebox{\linewidth}{!}{$
    \begin{bmatrix}
    1 & 0 & e_c \sin\theta_c\,dt & e_w \sin\theta_w\,dt & -e_w \cos\theta_w\,dt & e_c V_c \cos\theta_c\,dt & e_w (V_{lc}\sin\theta_w + V_{ld}\cos\theta_w)\,dt \\
    0 & 1 & e_c \cos\theta_c\,dt & e_w \cos\theta_w\,dt & e_w \sin\theta_w\,dt & -e_c V_c \sin\theta_c\,dt & e_w (-V_{ld}\sin\theta_w + V_{lc}\cos\theta_w)\,dt \\
    0 & 0 & 1 & 0 & 0 & 0 & 0 \\
    0 & 0 & 0 & 1 & 0 & 0 & 0 \\
    0 & 0 & 0 & 0 & 1 & 0 & 0 \\
    0 & 0 & 0 & 0 & 0 & 1 & 0 \\
    0 & 0 & 0 & 0 & 0 & 0 & 1
    \end{bmatrix},
    $}}
  \label{eq:F_jacobian_plus_full}
  \end{equation}
  \begin{equation}
    \mathbf{F}_k^{-} =
    \makebox[\linewidth][c]{
    \resizebox{\linewidth}{!}{$
    \begin{bmatrix}
    1 & 0 & e_c \sin\theta_c\,dt & e_w \sin\theta_w\,dt & e_w \cos\theta_w\,dt & e_c V_c \cos\theta_c\,dt & e_w (-V_{lc}\sin\theta_w + V_{ld}\cos\theta_w)\,dt \\
    0 & 1 & e_c \cos\theta_c\,dt & e_w \cos\theta_w\,dt & -e_w \sin\theta_w\,dt & -e_c V_c \sin\theta_c\,dt & e_w (-V_{ld}\sin\theta_w - V_{lc}\cos\theta_w)\,dt \\
    0 & 0 & 1 & 0 & 0 & 0 & 0 \\
    0 & 0 & 0 & 1 & 0 & 0 & 0 \\
    0 & 0 & 0 & 0 & 1 & 0 & 0 \\
    0 & 0 & 0 & 0 & 0 & 1 & 0 \\
    0 & 0 & 0 & 0 & 0 & 0 & 1
    \end{bmatrix},
    $}}
  \label{eq:F_jacobian_minus_full}
  \end{equation}
  with: 
  \begin{equation}
    e_c = \exp^{-\frac{\sigma_{\theta_c}}{2}}
  \end{equation}
  \begin{equation}
    e_w = \exp^{-\frac{\sigma_{\theta_w}}{2}}
  \end{equation}
}
This enables the filter to capture the weather effects on the covariance of the position of the missing person, even if the analytical solution of the speeds' standard deviation is not available.

The process noise for the propagation of the covariance is defined as:

{
\begin{equation}
  \mathbf{Q} = q \cdot \mathbf{I}_{7x7},
\end{equation}
where $q$ is a tunable parameter that scales the process noise of the filter. The tuning procedure for $q$ is outlined in section \ref{sec:results-tuning}.
}

The position of the person in the water is unknown throughout the search mission, but the weather information for the area is updated based on the measurements collected by the ship. The Jacobian of the measurement function is defined as:

\begin{equation}
  \mathbf{H}_k = \begin{bmatrix}
    \mathbf{0}_{5x2} &  & \mathbf{I}_{5x5}
  \end{bmatrix}.
  \label{eq:H_Jacobian}
\end{equation}
In a real application, the measurement noise will change according to the sensor specifications. For our simulations, the measurement noise is defined as:

{
\begin{equation}
  \mathbf{R} = r \cdot \mathbf{I}_{5x5},
  \label{eq:measurement_noise}
\end{equation}
where $r$ is a tunable parameter that scales the measurement noise of the filter. The full EKF algorithm is shown in Algorithm \ref{alg:ekf}. 
}

\begin{algorithm}[t]
  \caption{Extended Kalman Filter (EKF)}
  \begin{algorithmic}[1]
    \STATE Initialize the state estimate $\hat{\mathbf{x}}_0$ and covariance matrix $\mathbf{P}_0$.
    \FOR{$k = 1, 2, \dots$}
    \STATE \textbf{Predict Step:}

    \STATE $\hat{\mathbf{x}}_k = f(\hat{\mathbf{x}}_{k-1})$ \quad \text{(Nonlinear state transition)} \\
    \IF{$k\% 2 == 0$}
    \STATE $\mathbf{F}_{k-1} = \mathbf{F}_{k-1}^{+}$ \quad \text(Jacobian for positive crosswind)
    \ELSE
    \STATE $\mathbf{F}_{k-1} = \mathbf{F}_{k-1}^{-}$ \quad \text(Jacobian for negative crosswind)
    \ENDIF
    \STATE $\hat{\mathbf{P}_k} = \mathbf{F}_{k-1} \check{\mathbf{P}}_{k-1} \mathbf{F}_{k-1}^\top + \mathbf{Q}$ \quad \text{(Predicted covariance)} \\

    \STATE \textbf{Update Step:}
    \STATE $\mathbf{y}_k = \mathbf{z}_k - h(\check{\mathbf{x}}_k)$ \quad \text{(Innovation)} \\
    \STATE $\mathbf{S}_k = \mathbf{H}_k \check{\mathbf{P}}_k \mathbf{H}_k^\top + \mathbf{R}$ \quad \text{(Innovation covariance)} \\
    \STATE $\mathbf{K}_k = \check{\mathbf{P}}_k \mathbf{H}_k^\top \mathbf{S}_k^{-1}$ \quad \text{(Kalman gain)} \\
    \STATE $\hat{\mathbf{x}}_k = \check{\mathbf{x}}_k + \mathbf{K}_k \mathbf{y}_k$ \quad \text{(Updated state estimate)} \\
    \STATE $\hat{\mathbf{P}}_k = (\mathbf{I} - \mathbf{K}_k \mathbf{H}_k) \check{\mathbf{P}}_k$ \quad \text{(Updated covariance)}
    \ENDFOR
  \end{algorithmic}
  \label{alg:ekf}
\end{algorithm}

The estimation of the EKF is used to define a covariance ellipse, which constitutes the search area of the mission. The center of the ellipse is determined by the first two elements of the state vector $\mathbf{x}_k$, and the shape of the ellipse is determined by the eigenvalues and eigenvectors for the first two rows and columns of the covariance matrix $\mathbf{P}_k$. The lengths of the semi-major and semi-minor axes of the ellipse and their orientation are calculated by the eigenvalues and eigenvectors of the covariance matrix. The ellipse is scaled by a number of standard deviations $n_{std}$, based on the amount of information that needs to be covered.

\begin{table}[h]
  \centering
  \caption{Performance comparison of man-overboard prediction methods.}
  \begin{tabular}{|c|c|c|}
    \hline
    \textbf{Method}            & \textbf{Number of particles} & \textbf{Time (s)} \\ \hline
    Monte Carlo random particle    & 100               &  0.885           \\ \hline
    Monte Carlo random particle    & 1000              &  9.338           \\ \hline
    Monte Carlo random particle    & 10000             &  98.826           \\ \hline
    Extended Kalman Filter         & -                 &  \textbf{0.285} \\ \hline
  \end{tabular}
  \label{tab:performance-comparison}
\end{table}

A performance comparison between the Monte Carlo random particle model and the EKF method is shown in Table \ref{tab:performance-comparison}. The comparison is done for a $600$ second experiment. The time needed to execute the Monte Carlo random particle model is increasing exponentially with the number of particles used. The EKF method does not use sampling, so its more efficient. For each method, we show the average run time for $10$ executions. 

\section{Search methods}
\label{search_methods}
In this section we present the search methods that are employed to locate missing people at sea. The search methods explore the area defined by the covariance ellipse described in section \ref{ekf}. All the methods presented bellow are designed to work for dynamically expanding and moving areas.

After a person falls overboard the vessel, the bridge of the vessel is notified. Information about the last known position of the missing person and the weather are transmitted to the UAV, and the EKF method is initialized. The UAV estimates the search area and starts the search mission. The EKF updates the search area estimation based on the weather information. When a new area is estimated, new nodes are determined, based on the search pattern that is used for the search, and matched to the nodes of the previous time step. The mission terminates when the person is found, or when the UAV does not have sufficient battery to continue the mission and, therefore, returns to the vessel promptly and safely.

\begin{figure}[h]
  \centering
  \includegraphics[width=80mm]{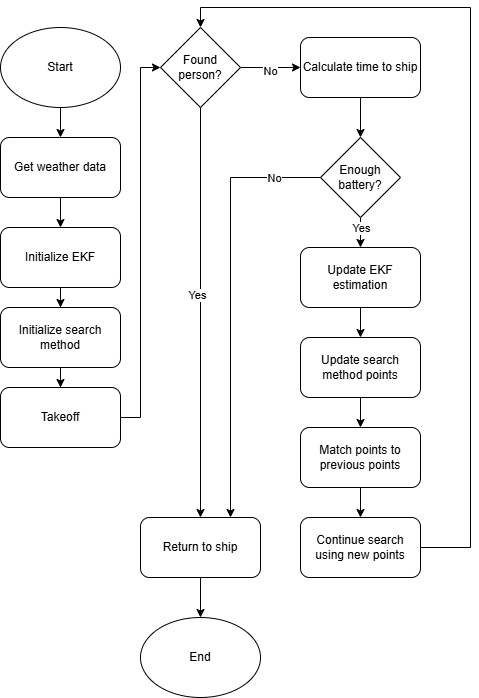}
  \caption{General search paradigm}
  \label{fig:general-search}
\end{figure}

The general framework of search is shown in Figure \ref{fig:general-search}. The search mission starts by the initialization of the weather parameters, EKF matrices and UAV parameters. Then, the EKF process is used to predict the search area. For every second after the UAV has initiated the search mission, the nodes of the current time step are calculated and matched to the nodes of the previous time step, and the UAV moves to the next nodes. The mission ends when the UAV has located the person, or it does not have sufficient battery to continue the mission and return to the ship safely.

\begin{figure}[h]
  \centering
  \begin{subfigure}[b]{0.4\textwidth}
    \centering
    \includegraphics[width=\textwidth]{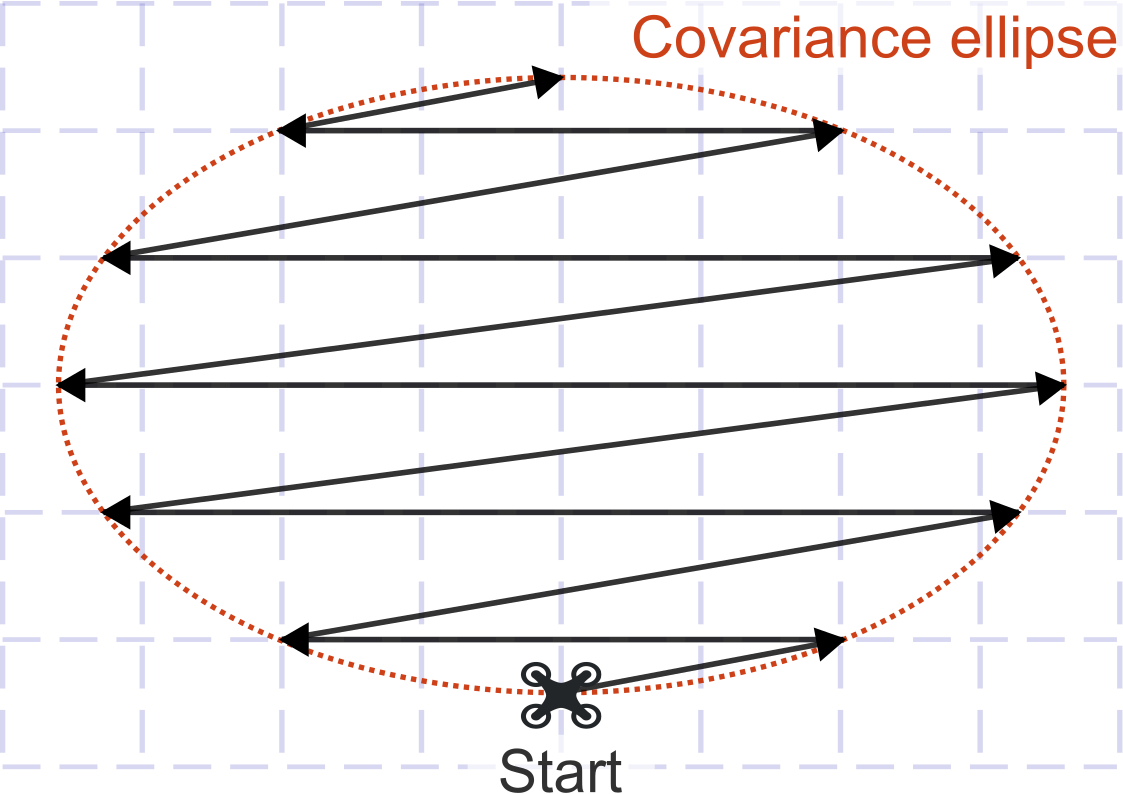}
    \caption{Zigzag pattern.}
    \label{fig:zigzag-pattern}
  \end{subfigure}
  \hfill
  \begin{subfigure}[b]{0.4\textwidth}
    \centering
    \includegraphics[width=\textwidth]{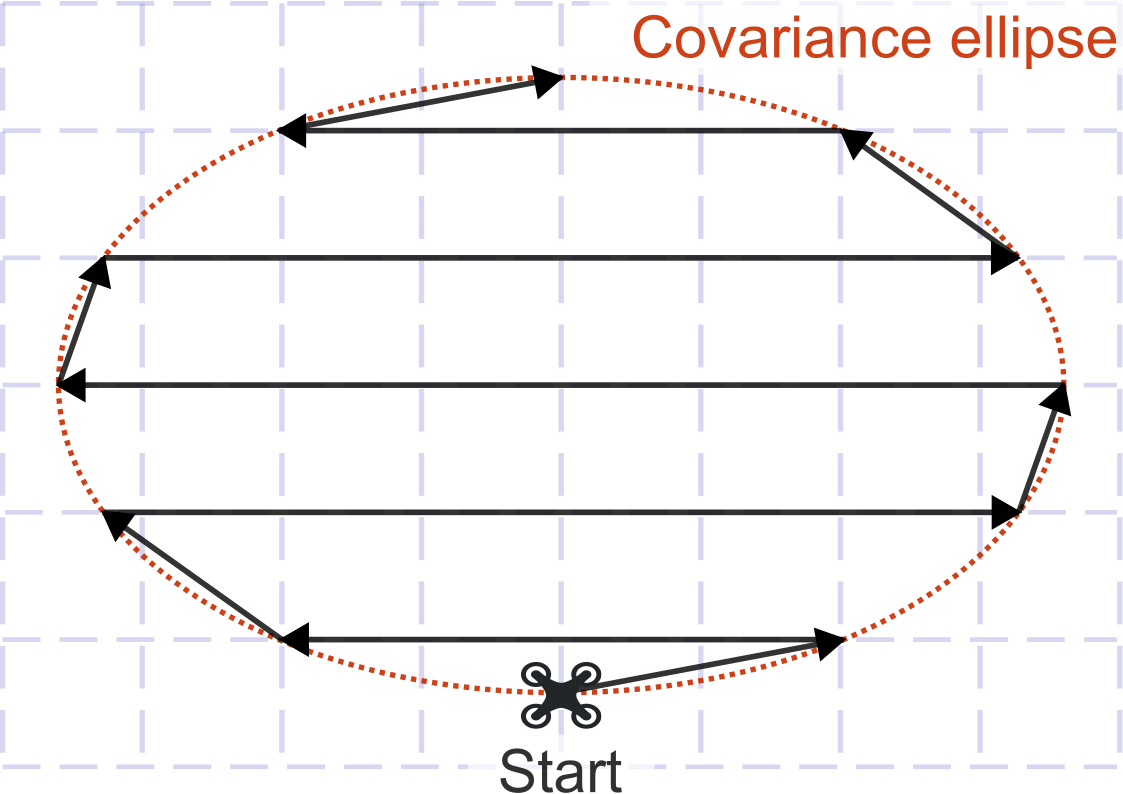}
    \caption{Boustrophedon pattern.}
    \label{fig:boustrophedon-pattern}
  \end{subfigure}
  \hfil
  \begin{subfigure}[b]{0.4\textwidth}
    \centering
    \includegraphics[width=\textwidth]{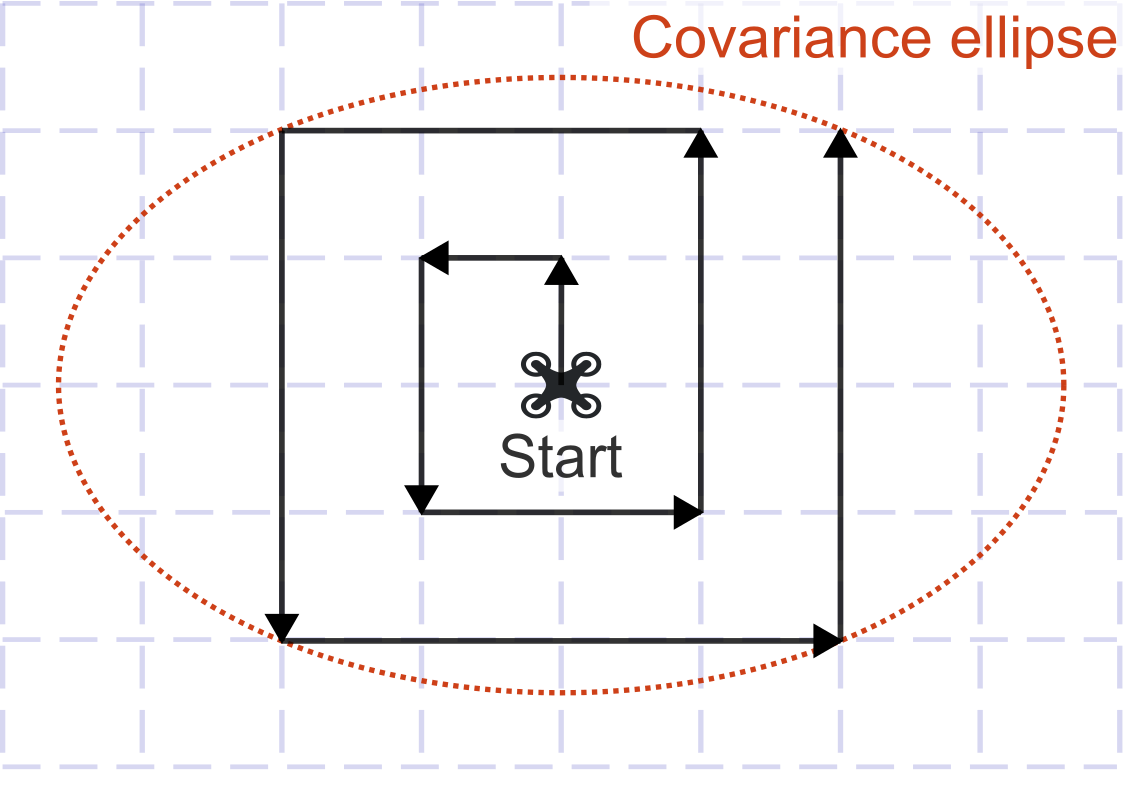}
    \caption{Spiral pattern.}
    \label{fig:spiral-pattern}
  \end{subfigure}

  \caption{Illustrations of naive search methods. The orange ellipse represents the predicted covariance ellipse and the black lines represent the paths for each search pattern.}
  \label{fig:main}
\end{figure}

\subsection{Zigzag}
The zigzag method is a systematic method for searching in an area. When the UAV starts a search mission, the node on the covariance ellipse that is the closest to its starting position is calculated. Starting from that node, parallel lines with width equal to the projected field of view of the camera of the UAV are drawn. In the cases where the UAV has more than one camera, the value of the one with the smallest field of view is used. The intersection nodes of the parallel lines and the covariance ellipse are the set of nodes the UAV has to traverse through. The UAV starts to go through the nodes starting from the start node and alternating between the right and left intersection nodes. The search area is updated every second. After the area is updated, the new nodes are calculated and matched with the previous nodes to ensure the smooth trajectory of the UAV. After the UAV has finished scanning the area, the method continues backwards, alternating every time it reaches the end. The zigzag method pattern can be seen in Figure \ref{fig:zigzag-pattern}.

\subsection{Boustrophedon}
The boustrophedon method is very similar to the zigzag method. The search procedure differs only on the sorting of the nodes in every iteration. After determining the start node, two nodes from each side of the starting node are selected instead of one, alternating between right to left nodes. This enables the method to fully cover the search area more times than the zigzag method. The boustrophedon method pattern can be seen in Figure \ref{fig:boustrophedon-pattern}.

\subsection{Spiral}
The spiral method is used by a single agent to completely cover the search area in an outward spiraling motion. After the UAV has initiated the search mission, we select the center of the estimated covariance ellipse as the center of the spiral. The set of nodes that the UAV has to traverse through are calculated by drawing imaginary horizontal and vertical lines with width equal to the projected field of view of the camera of the UAV, and identifying the ones that are included in the covariance ellipse area. Using polar coordinates centered at the ellipse center, the nodes are grouped based on their radial distance and sorted based on their angle, with angle zero on the x-axis. After the area is updated using this sorting, we match the new nodes to the previous nodes and the UAV moves towards the updated nodes. After the UAV has reached the end of the node set, it travels to the center of the ellipse and starts the procedure from the beginning. The spiral method pattern can be seen in Figure \ref{fig:spiral-pattern}.

\subsection{Probability informed search}
\label{sec:bayes}

The previous methods are used to fully cover the search area multiple times, without taking into account the density information from the covariance ellipse. This can be effective for small areas where multiple full passes are feasible, but it breaks down for bigger areas. {The probability informed search (PIS) utilizes the Bayesian Search Theory to search the areas where it is most probable to find people first.}

{
  The EKF provides on every time step $k$ a mean position $X_k \in \mathbb{R}^2$ and a covariance $P_K \in \mathbb{R}^{2 \times 2}$. The search domain is defined as the confidence ellipse: 
  \begin{equation}
    \Omega_k = \left\{x \in \mathbb{R}^2 : \left(x - X_k\right)^T P_k^{-1}\left(x-X_k\right) \leq \gamma\right\},
  \end{equation}
  where $\gamma$ determines the confidence level. The spatial support is directly induced by the EKF uncertainty and evolves at each time step. The ellipse is discretized into $N_k$ nodes, similar to the Spiral method: 
  \begin{equation}
    \mathcal{S}_k = {s_1, s_2, ..., s_{N_k}},
  \end{equation} 
  where each node represents a spatial cell of the ellipse on the current timestep. For each node, we initialize the prior from the EKF Gaussian as: 
  \begin{equation}
    \check{p}_{k} = \frac{\mathcal{N}(s_i; X_k,P_k)}{\sum_{j=1}^{N_k}\mathcal{N}(s_j; X_k, P_k)}.
  \end{equation}
}

{
  On each timestep $k$, the UAV observes a set of nodes in the ground projected field of view of its camera: 
  \begin{equation}
    \mathcal{F}_k \subseteq \mathcal{S}_k.
  \end{equation}
  We model the information gain of the UAV camera as a Bernoulli process: 
  \begin{equation}
    P\left(Z_k = 1 | X = s_i\right) = 
    \begin{cases}
      p_d, & s_i \in \mathcal{F}_k, \\
      0,   & s_i \notin \mathcal{F}_k.
    \end{cases}
  \end{equation}
  where $p_d$ represents the detection accuracy of the detection method used. For simplicity we assume that the method can not produce false positive cases.
  On every time step, we calculate the posterior probability of each visited node based on the Bayes rule:
  \begin{equation}
    \hat{p}_{k} (i) = \frac{P\left(z_k | X = s_i\right) \check{p}_{k} (i)}{\sum_{j=1}^{N_k} P\left(z_k | X = s_i\right)\check{p}_{k}(j)}.
  \end{equation}
  Applying the Bayes rule to update the probability of a node does not account for the diffusion of the probability for areas inside the ellipse that have been previously seen. To account for that, we calculate the size ratio between the current and previous predicted covariance ellipses and use it as a diffusion mechanism. On each timestep, we calculate the posterior probability of the unseen nodes as: 
  \begin{equation}
    \hat{p}_{k} (i) = \frac{\left(1 - P\left(z_k | X = s_i\right)\right) \check{p}_{k} (i)}{\sum_{j=1}^{N_k} \left(1-P\left(z_k | X = s_i\right)\right)\check{p}_{k}(j)} + r^{k}_{k-1} d_c, 
  \end{equation}
  where $r_{k-1}^{k}$ is the size ratio between the two ellipses between time steps $k$ and $k-1$, and $d_c$ is the dimension size of the projected field of view of the camera towards the direction of the neighboring node.
}

The method initializes when the UAV starts a search mission. The nodes that the UAV has to search are discretized similar to the spiral method. Each node is given a probability of containing a person based on the density
function of the covariance ellipse from the EKF, and all the nodes are sorted based on this probability. The UAV always heads towards the node with the highest probability of containing a person. When the UAV reaches a node, the Bayes rule is utilized to update the belief for that specific node and the next node is chosen. The diffusion mechanism is applied at every time step for a more accurate representation of the probability in the area.

This method utilizes the probability information from the EKF to choose areas with higher chances of containing people. Because the area is defined as a covariance ellipse, the nodes closest to the center of the ellipse have a much higher chance of containing people than the nodes closer to the ellipse boundaries. This results into the UAV moving erratically close to the center of the ellipse, passing nodes it has already covered multiple times. This is resolved with the improved probability informed search (IPIS) method in \ref{sec:astar_bayes}.

\begin{algorithm}[t!]
  \caption{A-star Search Algorithm}
  \begin{algorithmic}[1]
    \STATE openSet $\gets$ startNode \COMMENT{Nodes to explore}
    \STATE closedSet $\gets$ $\emptyset$ \COMMENT{Processed nodes}

    \WHILE{openSet is not empty}
    \STATE currentNode $\gets$ node with lowest $f$ in openSet
    \STATE closedSet.add(currentNode)
    \STATE openSet.remove(currentNode)

    \IF{currentNode $=$ goalNode}
    \RETURN ReconstructPath(currentNode)
    \ENDIF

    \FOR{n $\in$ currentNode.getNeighbors()}
    \IF{n $\in$ closedSet}
    \STATE \textbf{continue} \COMMENT{Skip processed nodes}
    \ENDIF

    \STATE f(n) $\gets$ g(currentNode) + h(n)d(n)
    \IF{n $\notin$ openSet or f(n) $<$ g(currentNode)}
    \STATE g(n) $\gets$ f(n)
    \STATE n.connection $\gets$ currentNode
    \IF{n $\notin$ openSet}
    \STATE openSet.add(n) \COMMENT{Add to open set if new or better path}
    \ENDIF
    \ENDIF
    \ENDFOR
    \ENDWHILE

    \STATE \RETURN $\emptyset$ \COMMENT{No path found}
  \end{algorithmic}
  \label{alg:astar}
\end{algorithm}

\subsection{Improved probability informed search}
\label{sec:astar_bayes}
The improved probability information search employs an A-star algorithm with an altered cost function to optimize the paths between nodes. The altered cost function is defined as:

\begin{equation}
  f(n) = g(n) + h(n)d(n),
\label{eq:improved-cost-function}
\end{equation}
where $f(n)$ is the total cost of the current node, $g(n)$ is the cost to go from the starting node to the current node, $h(n)$ is the estimated cost to go from the current node to the goal, and $d(n)$ is the probability of the person being in the area that the current node represents. The estimated cost to the goal is the Euclidean distance between the current node and the goal:

\begin{equation}
  h(n) = \sqrt{(x_n-x_{goal})^2 + (y_n-y_{goal})^2}
\end{equation}

The implementation of the A-star algorithm is given in Algorithm \ref{alg:astar}. We use the algorithm to calculate the best path from the current position to the node with the highest probability every time the EKF produces a new estimation area. This enables us to get a more optimal path than the method \ref{sec:bayes}, at the expense of extra computation time.

\section{Results}
In this section, we validate the performance of the EKF estimator that is described on section \ref{ekf}, and the performance of the search methods. To do so, we use a simulation environment that was developed for the purpose of this paper. All code for the simulation environment and the evaluation of the methods is available on our GitHub page at \url{https://github.com/diangeli/pdms-man-overboard}.

{\subsection{Ground truth comparison with OpenDrift}}

{
  OpenDrift is an open-source simulator for predicting how different objects in the ocean drift depending on the wind and current conditions of the simulated area. As mentioned in Section \ref{related-work}, it has been experimentally validated based on real life data. In this section, we perform an evaluation of our ground truth method by comparing its results to OpenDrift simulator to validate that our results are comparable and realistic.
}
  
{
  On Figure \ref{fig:drift_model_agreement}, we show all the necessary comparison data between the two simulators for one specific experiment. In Figures \ref{fig:drift_comparison_pointclouds} and \ref{fig:drift_comparison_density_contours} we show that the two distributions have small deviations, achieving mean distance difference of 18.4 meters and a symmetrical gaussian KL divergance score of 0.0154. From Figures \ref{fig:drift_comparison_radial_cdf} and \ref{fig:drift_comparison_radial_qq} we can conclude that the two simulators are in strong agreement at most of the range, with minor discrepancies at the extreme distances from the centers.
  \begin{figure}[h!]
    \centering
    \begin{subfigure}{0.48\linewidth}
      \centering
      \includegraphics[width=\linewidth]{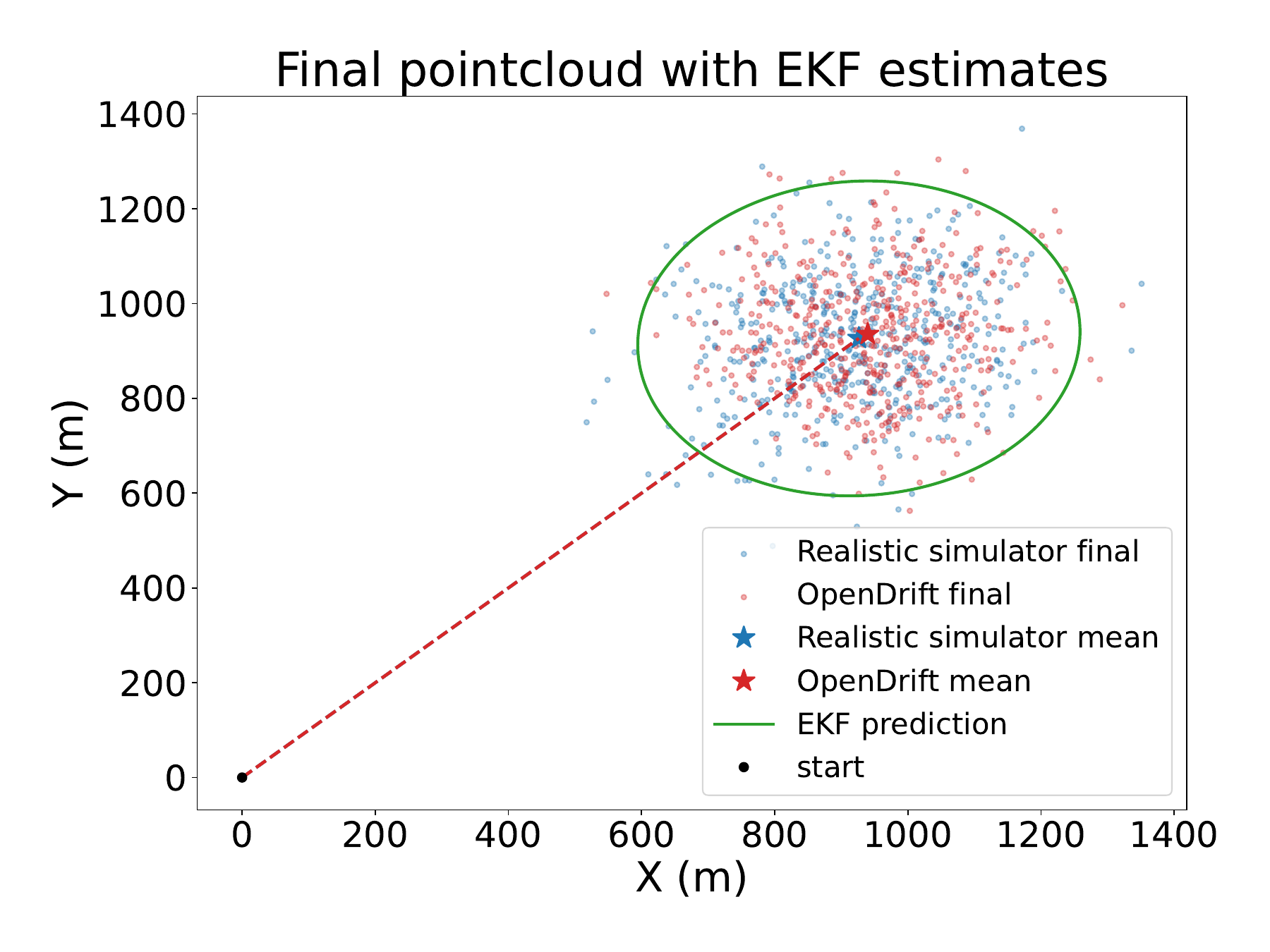}
      \caption{Final position of pointclouds for our realistic simulator and OpenDrift simulator. In green we show the EKF prediction for this experiment.}
      \label{fig:drift_comparison_pointclouds}
    \end{subfigure}
    \hfill
    \begin{subfigure}{0.48\linewidth}
      \centering
      \includegraphics[width=\linewidth]{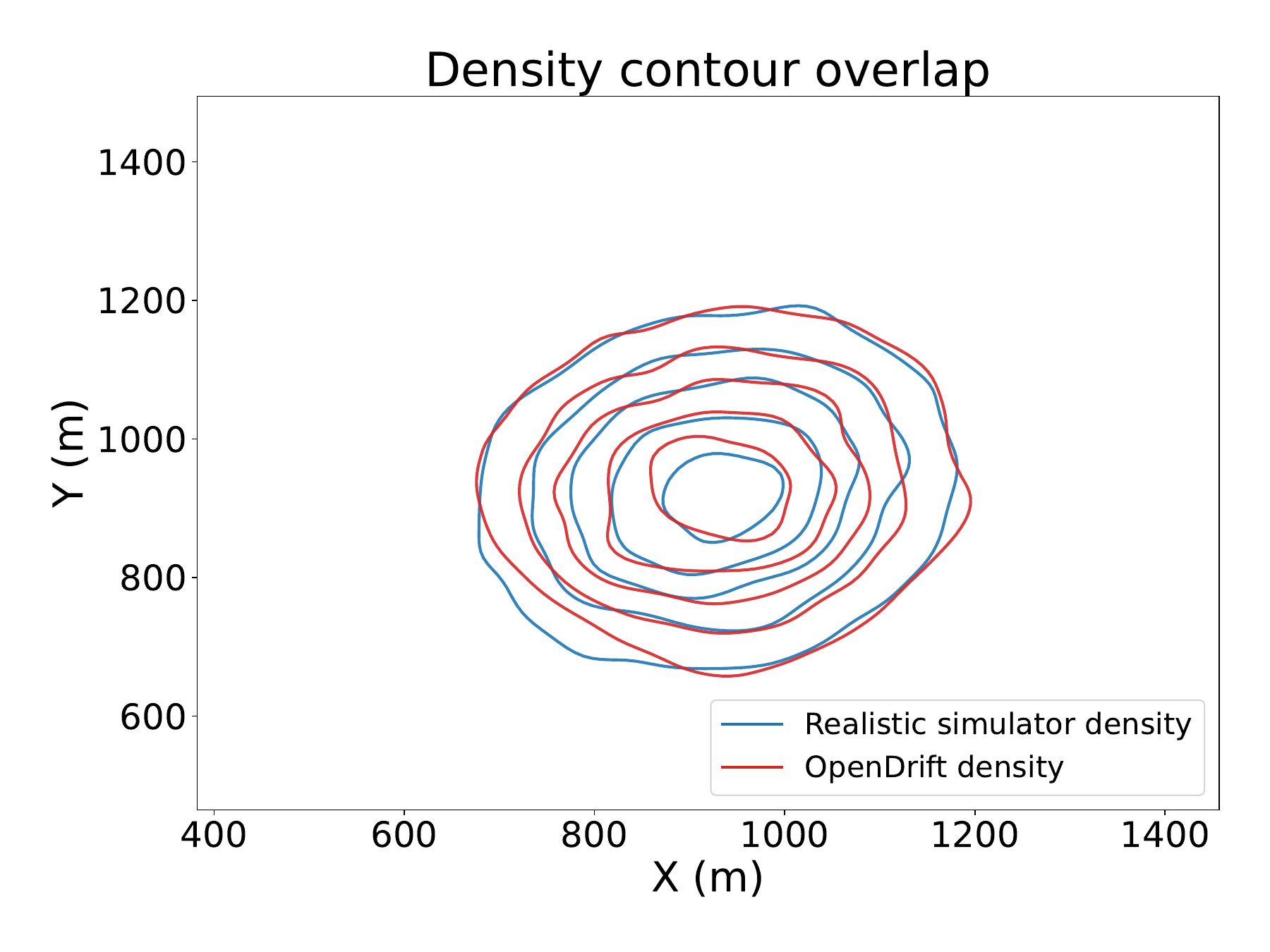}
      \caption{Overlay of 2D density contours of final particle positions for both our realistic simulator and OpenDrift.}
      \label{fig:drift_comparison_density_contours}
    \end{subfigure}
    \vspace{0.5cm}
    \begin{subfigure}{0.48\linewidth}
      \centering
      \includegraphics[width=\linewidth]{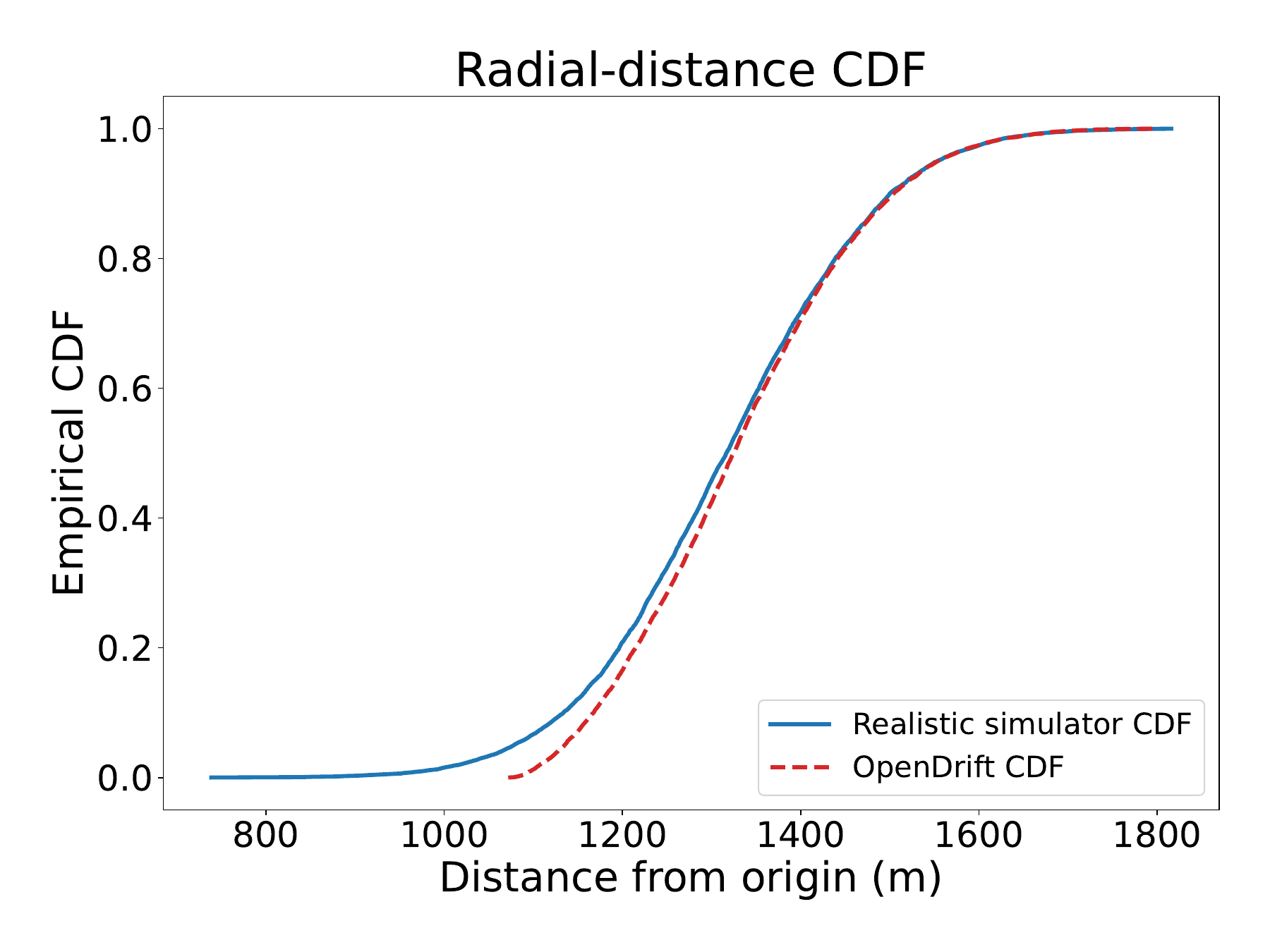}
      \caption{Empirical CDF of radial distance from the origin for realistic simulator and OpenDrift.}
      \label{fig:drift_comparison_radial_cdf}
    \end{subfigure}
    \hfill
    \begin{subfigure}{0.48\linewidth}
      \centering
      \includegraphics[width=\linewidth]{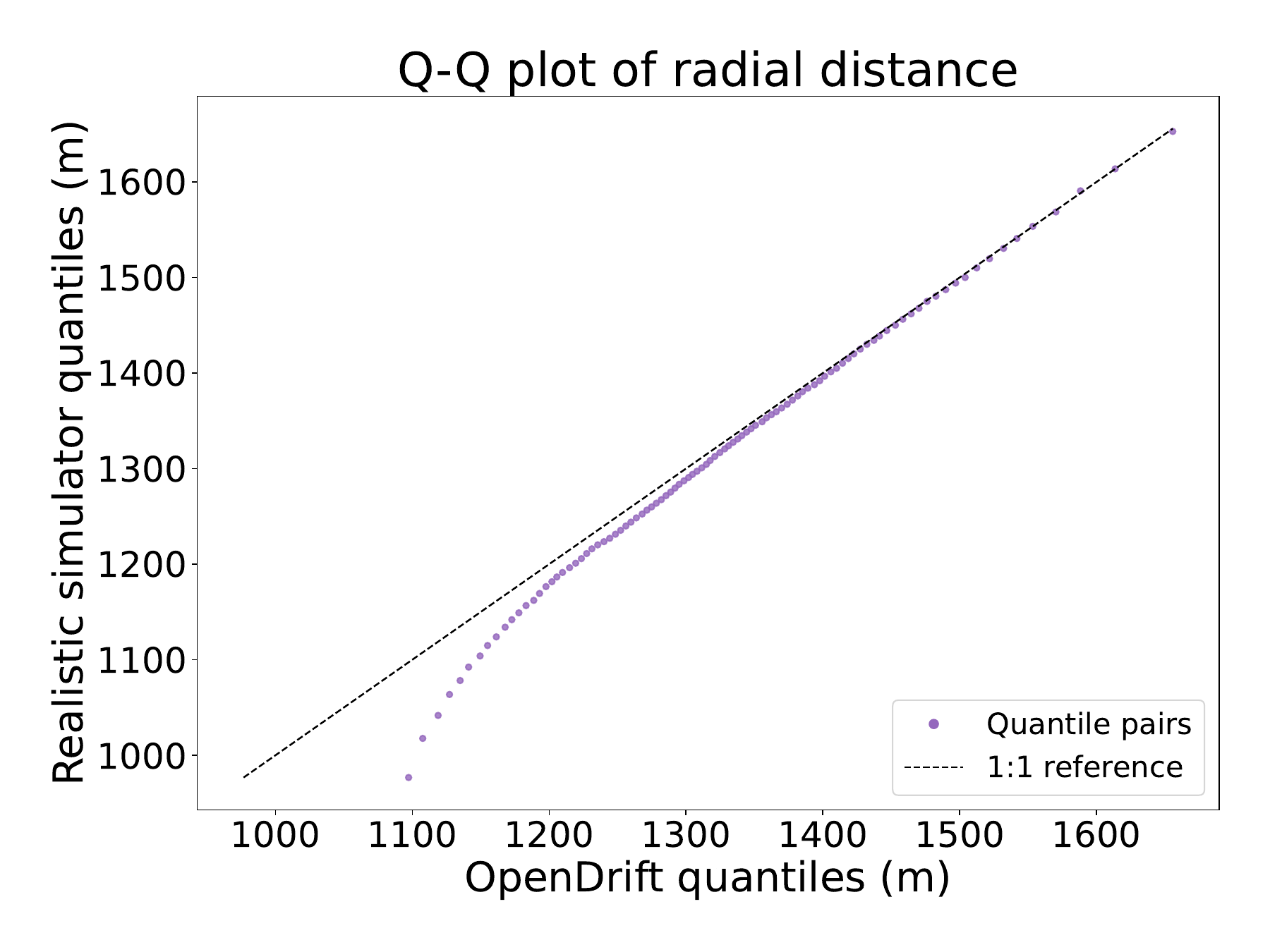}
      \caption{Q--Q plot of radial distance quantiles for realistic simulator vs OpenDrift.}
      \label{fig:drift_comparison_radial_qq}
    \end{subfigure}
    \caption{Comparison of final-state drift distributions produced by our realistic simulator (blue) and OpenDrift simulator (red) for the same simulation settings.}
    \label{fig:drift_model_agreement}
  \end{figure}
}

\subsection{EKF tuning and performance}
\label{sec:results-tuning}
Here, we tune the EKF estimator described on section \ref{ekf}. The purpose of the tuning is to enable the EKF to produce predictions that closely follow the distribution of the particles that are produced from the Monte Carlo random particle method. The prediction should overestimate the area, so that all or most of the particles are contained. We scale the covariance ellipse with 4 standard deviations, which results in $99.9937\%$ of the particles to be included in the area.

\begin{table}[t]
  \centering
  \caption{Weather Parameter Values for Tuning and Testing}
  \begin{tabular}{|c|c|c|}
    \hline
    \textbf{Variable}            & \textbf{Tuning} & \textbf{Evaluation} \\ \hline
    $\mu_{V_c} (\frac{m}{s})$    & 1               & $[1, 2]$            \\ \hline
    $\sigma_{V_c} (\frac{m}{s})$ & 0.2             & $[0.1, 0.2]$        \\ \hline
    $\mu_{V_w}(\frac{m}{s})$     & {15} & {$[10, 20]$}          \\ \hline
    $\sigma_{V_w} (\frac{m}{s})$ & 5               & $[2, 5]$            \\ \hline
    $\mu_{\theta_c}(\degree)$    & 0               & $[0]$               \\ \hline
    $\sigma_{\theta_c}(\degree)$ & 5               & $[0, 5, 10]$        \\ \hline
    $\mu_{\theta_w}(\degree)$    & 90              & $(0, 360, 45)$      \\ \hline
    $\sigma_{\theta_w}(\degree)$ & 5               & $[0, 5, 10]$        \\ \hline
  \end{tabular}
  \label{tab:tuning-parameters}
\end{table}

{
  We tune the filter by testing the different $q$ and $r$ values in one experiment setup. The wind and current data are representative of realistic conditions found at the ocean. We choose moderate uncertainty values for the weather measurements, so that the filter can model realistic conditions. The tuning parameters that were chosen for the tuning can be seen in Table \ref{tab:tuning-parameters}. We evaluate the results based on the Kullback-Leibler (KL) divergence between the ground truth distribution and the EKF prediction on the last time step of each experiment.
}

{The tuning procedure produced optimal values $q = 4.0$ and $r = 1.0$.} We validate the performance of the EKF by running experiments with the parameters shown in Table \ref{tab:tuning-parameters}. This set of parameters contains a collection of realistic scenarios, that provide a good representation of the performance of the filter in conditions that are encountered at the oceans.

{
  \begin{table}[H]
  \centering
  \caption{Evaluation of tuning}
  \begin{tabular}{|l|l|l|}
    \hline
                                & \textbf{RMSE} & \textbf{KL} \\ \hline
    \textbf{Mean}               & 1.9174        & 0.4659      \\ \hline
    \textbf{Standard Deviation} & 1.3536        & 0.5527      \\ \hline
  \end{tabular}
  \label{tab:tuning-evaluation}
\end{table}
}

In Table \ref{tab:tuning-evaluation} we present the results from the evaluation of the filter. Overall, the filter prediction represents the ground truth data distribution accurately. The low RMSE value indicates that the filter tracks the mean of the ground truth data very closely, which is expected since the state transition matrix contains the analytical solution of the mean speed of the particles. The KL divergence score is also very low showing that the predicted covariance ellipse represents the ground truth ellipse accurately.

{
  \subsection{EKF sensitivity analysis}
  In this section, we conduct a sensitivity analysis for the noise parameters of the EKF. Since the performance of the EKF is strongly influenced by the assumed process and measurement noise statistics, the purpose of this sensitivity analysis is to evaluate the robustness of the estimator with respect to variations in the noise covariance matrices. 
  }

\begin{figure}[h]
  \centering
  \includegraphics[width=135mm]{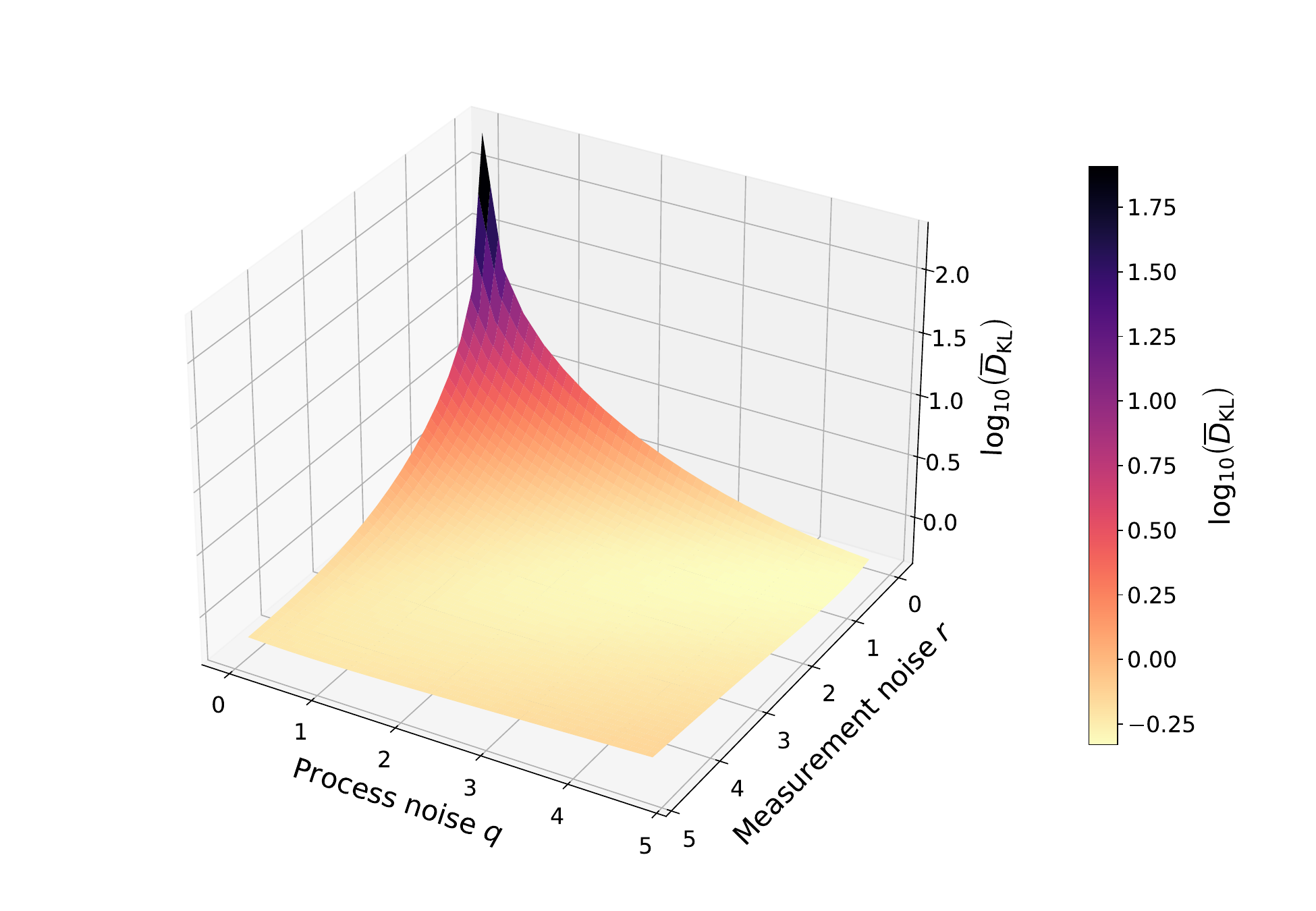}
  \caption{{Sensitivity of EKF performance to process and measurement noise covariance scaling. The color and surface height represent the logarithm of the mean KL-divergence between the true and estimated distributions as a function of the process scaling factor $q$ and measurement noise scaling factor $r$.}}
  \label{fig:EKF-sensitivity}
\end{figure}

{
  Figure \ref{fig:EKF-sensitivity} shows the surface of $\log_{10}\left(\overline{D_{KL}}\right)$ over the process noise scaling $q$ and measurement noise scaling $r$. A pronounced peak appears at very small values for q and r, indicating substantially increase divergence between the ground truth particle distribution and the EKF distribution. As either parameter becomes bigger, the divergence drops drastically and transitions into an almost flat region for moderate to high values, suggesting low sensitivity to variations. The selected tuning values $q=4.0$ and $r=1.0$ lie within this low-divergence area.
}

\subsection{Performance of the A-star probability informed cost function}

In this section, we discuss the performance of the improved A-star cost function described in section \ref{sec:astar_bayes}. This cost function incorporates the probability of a person being at a specific node in the calculation of the optimal detection path for the IPIS method. We perform experiments for full coverage of a static area to showcase the prowess of the method. We compare the improved cost function to the original A-star cost function.

\begin{figure}[h]
  \centering
  \includegraphics[width=135mm]{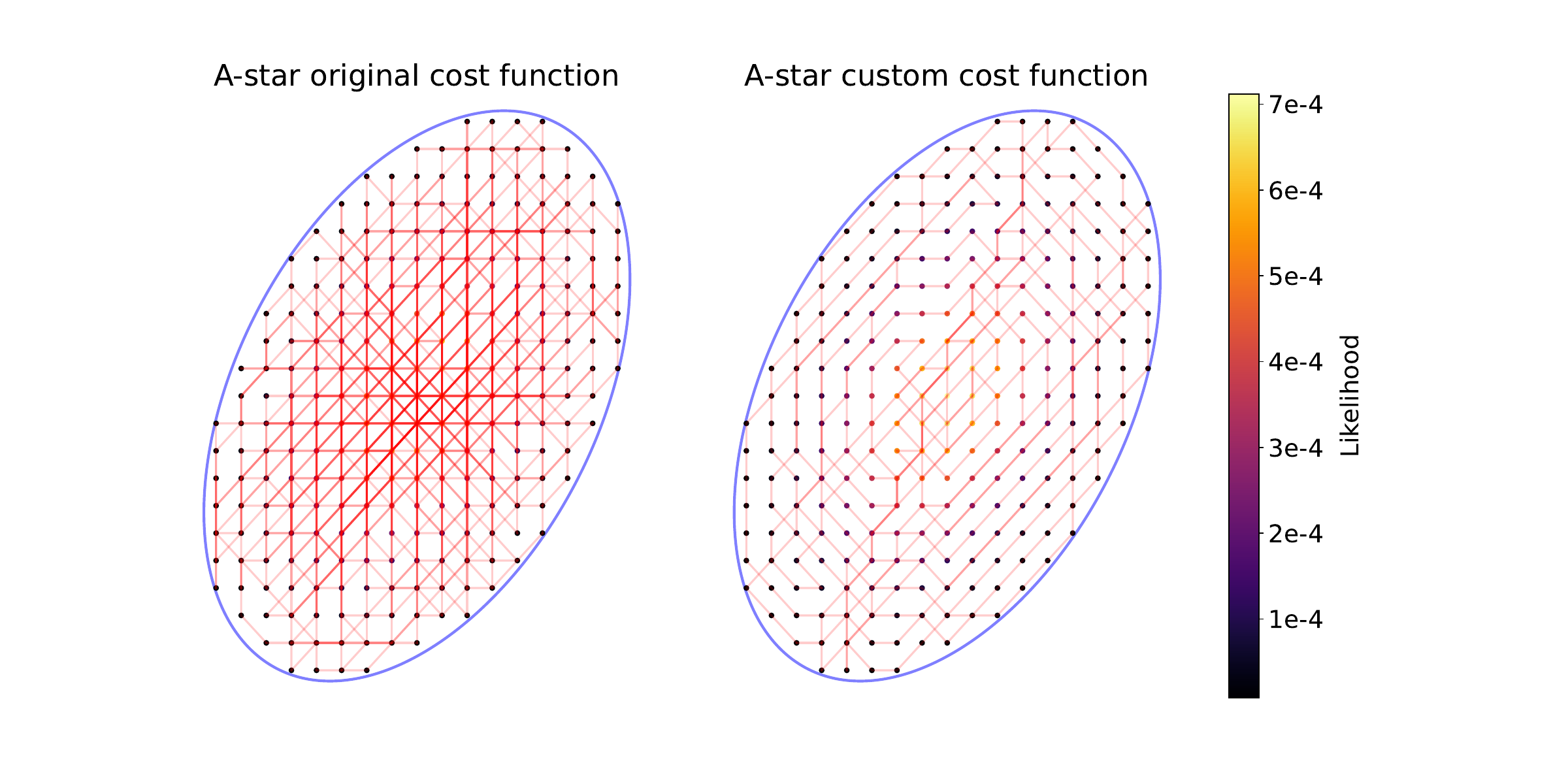}
  \caption{Performance comparison of base and custom A-star cost functions on static areas for full coverage. The ellipse represents the plotted covariance ellipse for the static area experiment. The color of each node shows the likelihood of a person being in the field of view of the camera when the UAV is at that position. The straight lines show the paths that were created during the method. The less transparent a line is, the less it was used during the path planning.}
  \label{fig:cost-function-comparison}
\end{figure}

In Figure \ref{fig:cost-function-comparison}, we show the performance comparison of the original cost function and the custom cost function for the full coverage of a static covariance ellipse area. The experiment was performed for a covariance ellipse that is defined as:

\begin{equation}
  \{ x:(x-\mu)^{T} \Sigma^{-1} (x-\mu) = 3\},
\end{equation}
where
\begin{equation}
  \boldsymbol\mu = \begin{bmatrix}0\\0\end{bmatrix}
  \quad\text{and}\quad
  \boldsymbol\Sigma = \begin{bmatrix}
  100 & 200\\
  200 & 300
  \end{bmatrix}.
\end{equation}
We create nodes inside the ellipse with a node distance of $5$ meters, as described in section \ref{sec:bayes}.

We show that the improved cost function enables the IPIS method to fully cover the area faster. The custom cost function forces the UAV to choose visiting nodes that have not been visited yet, limiting the number of multiple visits to the same nodes. This enables the method to produce paths that have not been traveled, increasing the chance of finding the missing person.

\subsection{Quantitative comparison of search methods}
\label{sec:quantitative-results}

In this section, we discuss the performance of the search methods presented on section \ref{search_methods}. The evaluation is based on quantitative data that were collected based on multiple experiments for all methods, under diverse conditions.

{
  For this set of experiments, the UAV moves with an average speed of $20$ meters per second and has a battery life of $30$ minutes. These parameters were selected based on the maritime SAR UAV that is presented in \cite{dimos}, but can also be achieved by a number of commercial UAVs \cite{uav1,uav2,uav3,uav4}. Every second of the simulation we calculate if the UAV would be able to return to the ship based on the average speed and the remaining battery life. When the battery life is not sufficient for the return trip, the experiment is terminated. The simulated camera collects images at $30fps$, and has a field of view of $45$ by $45$ degrees. The camera is simulated as a static camera on the UAV frame, with a downward orientation. For the ground truth, we initialize the Monte Carlo random particle method with 100 particles, each representing a person. The rest of the experiment parameters are shown in Table \ref{tab:testing-parameters}. We perform experiments for all different combinations of these parameters.
}

We assume that the detection method is able to detect people that appear in the field of view of the camera with a $30\%, 50\%$ and $100\%$ accuracy. For example, a detection method with a $30\%$ accuracy can find a person in a frame with a $30\%$ chance. This enables us to showcase the performance of the search methods for a diverse quality of detection methods. {As mentioned in Section \ref{sec:bayes}, we assume that the detection method does not produce false positive of negative predictions. }

{
  By combining all the different scenarios, we ran 69,120 experiments for each search and detection method combination. This brings the total number of experiments to 1,036,800. The number of experiments emphasizes the importance of conducting the analysis in simulation, to account for a multitude of edge cases not often observed in real tests. The results of the simulation experiments are shown in Tables \ref{tab:results}, \ref{tab:results_wind} and \ref{tab:results_current}. 
}

\begin{table}[t]
  \centering
  \caption{Environmental and UAV Parameter Values for Tuning and Testing}
  \begin{tabular}{|c|c|}
    \hline
    \textbf{Variable} & \textbf{Evaluation} \\ \hline
    $V_{w}\,(\mathrm{m/s})$ & {[5, 10, 15, 20, 25]} \\ \hline
    $\sigma_{V_w}\,(\mathrm{m/s})$ & [5, 10] \\ \hline
    $\theta_{w}\,(\degree)$ & [0, 45, 90, 135, 180, 225, 270, 315] \\ \hline
    $\sigma_{\theta_w}\,(\degree)$ & [5, 10, 15] \\ \hline
    $V_{c}\,(\mathrm{m/s})$ & [0.5, 1, 1.5, 2] \\ \hline
    $\sigma_{V_c}\,(\mathrm{m/s})$ & [0.25, 0.5] \\ \hline
    $\sigma_{\theta_c}\,(\degree)$ & [5, 10, 15] \\ \hline
    $t_{\mathrm{start}}^{\mathrm{UAV}}\,(\mathrm{s})$ & [300, 600, 900, 1200] \\ \hline
    $h_{\mathrm{UAV}}\,(\mathrm{m})$ & {[60, 90, 120]} \\ \hline
  \end{tabular}
  \label{tab:testing-parameters}
\end{table}

Firstly, the search methods are evaluated based on the average success rate of the missions. The average success rate is calculated by averaging the number of particles each method was able to identify at the end of each experiment. {Secondly, the search methods are compared based on how fast the UAV can find $50\%$ of the people in the area ($F_{50}$).} Fast missing person identification is directly correlated with a higher probability of survival.

Both of these metrics are combined by calculating the Mean Area Under the Curve (MAUC) metric. MAUC is determined by calculating the area under each search method's curve in Figures \ref{fig:performance100}, \ref{fig:performance50}, and \ref{fig:performance30}. A high MAUC value essentially means that the search method finds more people faster, while also producing a high success rate. In order to match data collected from different experiments, the X-axis represents the time of flight of the UAV. To facilitate the MAUC comparison among all methods, we pad the results of all the experiments to match the total time length of the slowest one. This can be seen in Figure \ref{fig:performancombined}.

{
  On Table \ref{tab:results}, we show the performance of each of the method for the different detection thresholds. The IPIS method outperforms all other methods in most cases. In the case of the perfect detection method, we observe that the Spiral method produces higher average success rate, but does so slower than IPIS. The Zigzag and Boustrophedon methods perform the worst in all cases. These methods do not start searching from the high probability areas, aka the center of the predicted ellipse, producing sub-par results compared to the other methods. Overall, the IPIS method is the better method to use on average.
}

{
  \begin{table}[h]
  \caption{Simulation results based on detection thresholds}
  \centering
  \begin{NiceTabular}{*{5}{c}}[hvlines]
    DA (\%)          & Method        & Success (\%)   & $F_{50}$ (s) & MAUC              \\
    \Block{5-1}{0.3} & Zigzag        & 51.80          & 656.81       & 24967.92          \\
    ~                & Boustrophedon & 49.15          & 695.44       & 22660.39          \\
    ~                & Spiral        & 77.03          & 266.69       & 56431.37          \\
    ~                & PIS           & 69.09          & 338.32       & 54380.15          \\
    ~                & IPIS          & \textbf{80.40} & \textbf{230.42} & \textbf{62450.05} \\ \hline

    \Block{5-1}{0.5} & Zigzag        & 58.54          & 630.92       & 28247.53          \\
    ~                & Boustrophedon & 54.56          & 675.03       & 25308.43          \\
    ~                & Spiral        & \textbf{82.91}          & 238.69       & 61239.07          \\
    ~                & PIS           & 70.36          & 319.69       & 55980.86          \\
    ~                & IPIS          & 81.86 & \textbf{206.91} & \textbf{64694.06} \\ \hline

    \Block{5-1}{1.0} & Zigzag        & 61.13          & 622.00       & 29550.31          \\
    ~                & Boustrophedon & 56.04          & 669.02       & 26077.99          \\
    ~                & Spiral        & \textbf{85.09} & 226.44       & 63181.46          \\
    ~                & PIS           & 71.24          & 306.93       & 57044.86          \\
    ~                & IPIS          & 82.57          & \textbf{194.62} & \textbf{65797.97} \\
  \end{NiceTabular}
  \label{tab:results}
\end{table}
}

{
  On Table \ref{tab:results_wind}, we show the average performance of the methods under varying wind speeds. At lower wind conditions, the lower particle dispersion favors probabilistic approaches that focus on the high density areas first, with the IPIS method achieving the highest success rates and MAUC values together with the lowest $F_{50}$. As wind speed increases, overall performance degrades for all methods. Spiral method retains a comparatively higher success rate at the maximum tested wind speed (25 m/s), outperforming the remaining methods in detection rate. In contrast, Zigzag and Boustrophedon consistently yield the lowest success rates and MAUC across all wind conditions, indicating limited robustness to wind-induced dispersion.
}

{
  \begin{table}[H]
  \caption{Simulation results based on wind speeds}
  \centering
  \begin{NiceTabular}{*{5}{c}}[hvlines]
    Wind (m/s)       & Method        & Success (\%)   & $F_{50}$ (s) & MAUC              \\

    \Block{5-1}{5}  & Zigzag        & 58.48          & 621.93       & 26817.32          \\
    ~                & Boustrophedon & 52.71          & 675.89       & 23060.00          \\
    ~                & Spiral        & 90.81          & 158.44       & 69188.78          \\
    ~                & PIS           & 91.80          & 85.33        & 74151.74          \\
    ~                & IPIS          & \textbf{97.07} & \textbf{71.70} & \textbf{78592.17} \\ \hline

    \Block{5-1}{10} & Zigzag        & 58.04          & 628.49       & 27236.55          \\
    ~                & Boustrophedon & 53.27          & 677.28       & 23861.69          \\
    ~                & Spiral        & 87.86          & 188.95       & 65277.98          \\
    ~                & PIS           & 80.93          & 181.80       & 64373.23          \\
    ~                & IPIS          & \textbf{90.82} & \textbf{106.99} & \textbf{71965.56} \\ \hline

    \Block{5-1}{15} & Zigzag        & 57.39          & 635.64       & 27655.45          \\
    ~                & Boustrophedon & 53.53          & 679.28       & 24712.91          \\
    ~                & Spiral        & \textbf{83.05} & 228.01       & 60813.19          \\
    ~                & PIS           & 69.38          & 318.62       & 54781.77          \\
    ~                & IPIS          & 82.29          & \textbf{186.92} & \textbf{64325.62} \\ \hline

    \Block{5-1}{20} & Zigzag        & 56.54          & 643.54       & 28039.26          \\
    ~                & Boustrophedon & 53.52          & 681.72       & 25538.43          \\
    ~                & Spiral        & \textbf{76.68} & \textbf{287.35} & 55554.24          \\
    ~                & PIS           & 58.97          & 463.54       & 46395.33          \\
    ~                & IPIS          & 73.30          & 290.10       & \textbf{56825.48} \\ \hline

    \Block{5-1}{25} & Zigzag        & 55.34          & 653.31       & 28194.34          \\
    ~                & Boustrophedon & 53.21          & 684.97       & 26238.33          \\
    ~                & Spiral        & \textbf{69.98} & \textbf{356.97} & \textbf{50585.65} \\
    ~                & PIS           & 50.08          & 558.94       & 39307.71          \\
    ~                & IPIS          & 64.59          & 397.54       & 49861.30          \\

  \end{NiceTabular}
  \label{tab:results_wind}
\end{table}
}

{
  On Table \ref{tab:results_current}, we show the average performance of all methods under different current speeds. At low current speeds, the particles move less, allowing structured patterns such as Spiral to achieve higher success rates. As the current speed increases, the particles move with higher speeds, and the methods exploiting the probability information of the EKF perform better. In particular, IPIS consistently attains the highest success rate and MAUC for current speeds above 1.0 m/s, while maintaining the lowest $F_{50}$, indicating superior robustness under strong current conditions.
}

{
  \begin{table}[H]
  \caption{Simulation results based on current speeds}
  \centering
  \begin{NiceTabular}{*{5}{c}}[hvlines]
    Current (m/s)    & Method        & Success (\%)   & $F_{50}$ (s) & MAUC              \\

    \Block{5-1}{0.5} & Zigzag        & 67.00          & 577.90       & 32539.13          \\
    ~                & Boustrophedon & 67.14          & 599.33       & 31702.06          \\
    ~                & Spiral        & \textbf{85.46} & \textbf{153.57} & 67361.97      \\
    ~                & PIS           & 71.78          & 288.17       & 58791.06          \\
    ~                & IPIS          & 83.09          & 169.29       & \textbf{67840.52} \\ \hline

    \Block{5-1}{1.0} & Zigzag        & 62.27          & 604.43       & 30515.47          \\
    ~                & Boustrophedon & 60.19          & 637.49       & 28330.23          \\
    ~                & Spiral        & \textbf{84.42} & \textbf{195.55} & 64317.91      \\
    ~                & PIS           & 70.21          & 315.71       & 56543.02          \\
    ~                & IPIS          & 82.17          & 197.18       & \textbf{65512.62} \\ \hline

    \Block{5-1}{1.5} & Zigzag        & 54.15          & 654.83       & 25990.79          \\
    ~                & Boustrophedon & 48.67          & 706.09       & 22395.58          \\
    ~                & Spiral        & 80.09          & 282.82       & 57076.60          \\
    ~                & PIS           & 70.14          & 328.46       & 55101.07          \\
    ~                & IPIS          & \textbf{81.13} & \textbf{225.29} & \textbf{63159.55} \\ \hline

    \Block{5-1}{2.0} & Zigzag        & 45.20          & 709.16       & 21308.95          \\
    ~                & Boustrophedon & 36.99          & 776.41       & 16301.22          \\
    ~                & Spiral        & 76.74          & 343.82       & 52379.41          \\
    ~                & PIS           & 68.79          & 354.24       & 52772.68          \\
    ~                & IPIS          & \textbf{80.06} & \textbf{250.84} & \textbf{60743.42} \\

  \end{NiceTabular}
  \label{tab:results_current}
\end{table}
}

These results can also be validated in Figure \ref{fig:performancombined}. Here, we visualize the average number of people saved per second for each of the search methods. The Spiral, PIS and IPIS methods are much faster than the rest in finding the missing people. {As the detection method performance declines, so does the overall performance of all methods. Nevertheless, the trends stay consistent, with the IPIS method outperforming all other methods for the lowest performing detection method.} 

{
  \begin{figure}[p]
  \centering
  \begin{subfigure}{\linewidth}
    \centering
    \caption{Performance of search methods for detection accuracy of $100\%$.}
    \includegraphics[width=135mm]{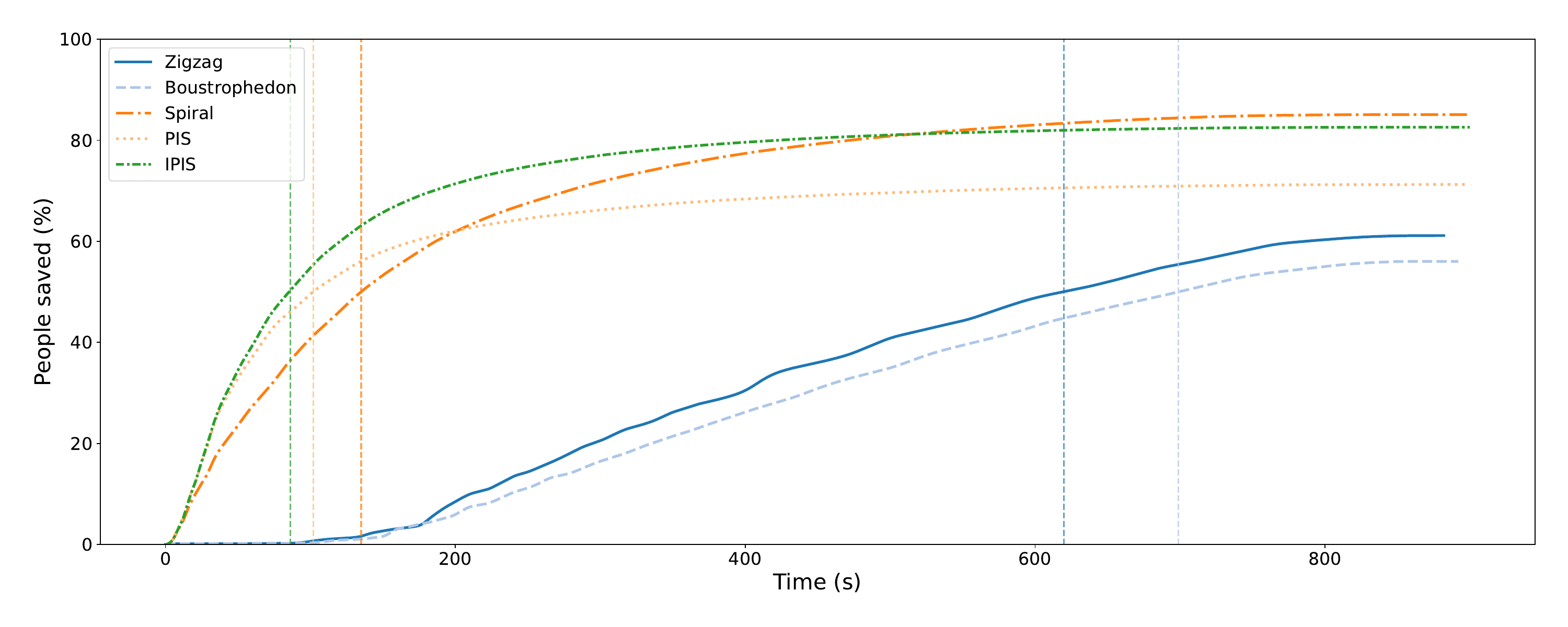}
    \label{fig:performance100}
  \end{subfigure}
  \begin{subfigure}{\linewidth}
    \centering
    \caption{Performance of search methods for detection accuracy of $50\%$.}
    \includegraphics[width=135mm]{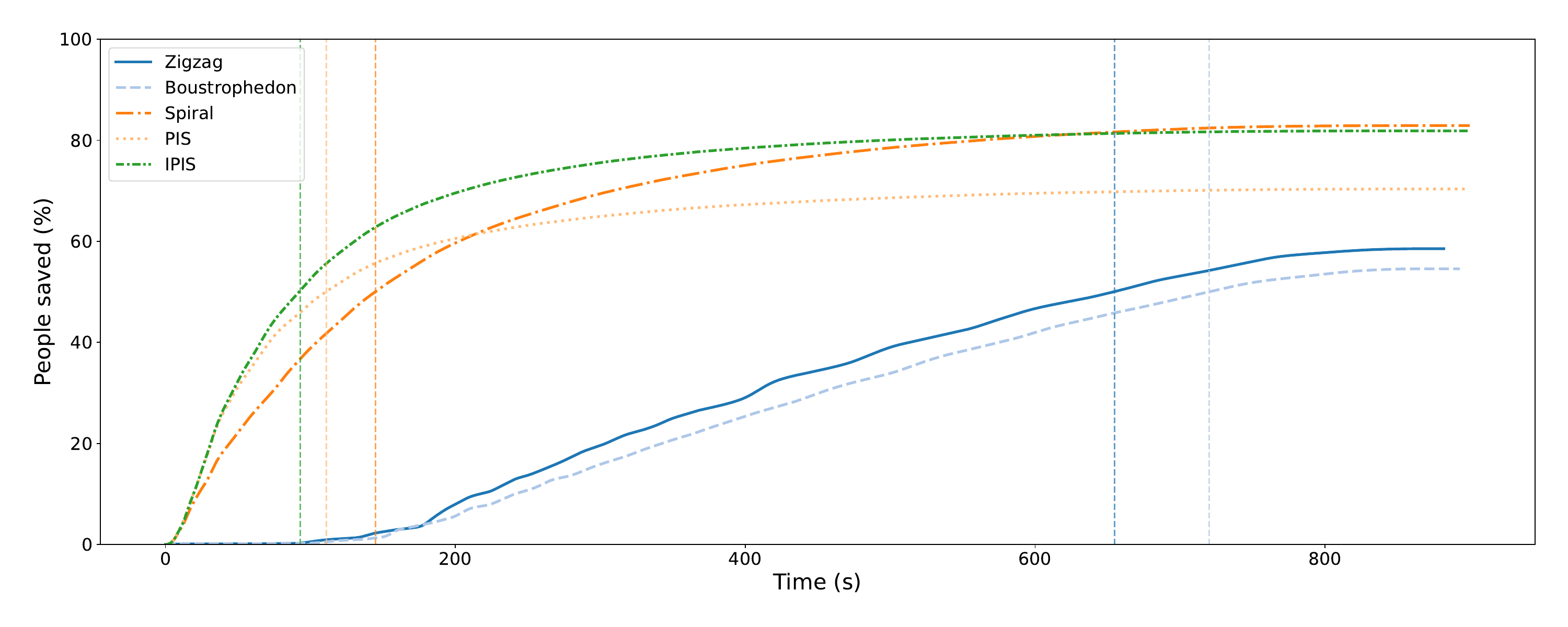}
    \label{fig:performance50}
  \end{subfigure}
  \begin{subfigure}{\linewidth}
    \centering
    \caption{Performance of search methods for detection accuracy of $30\%$.}
    \includegraphics[width=135mm]{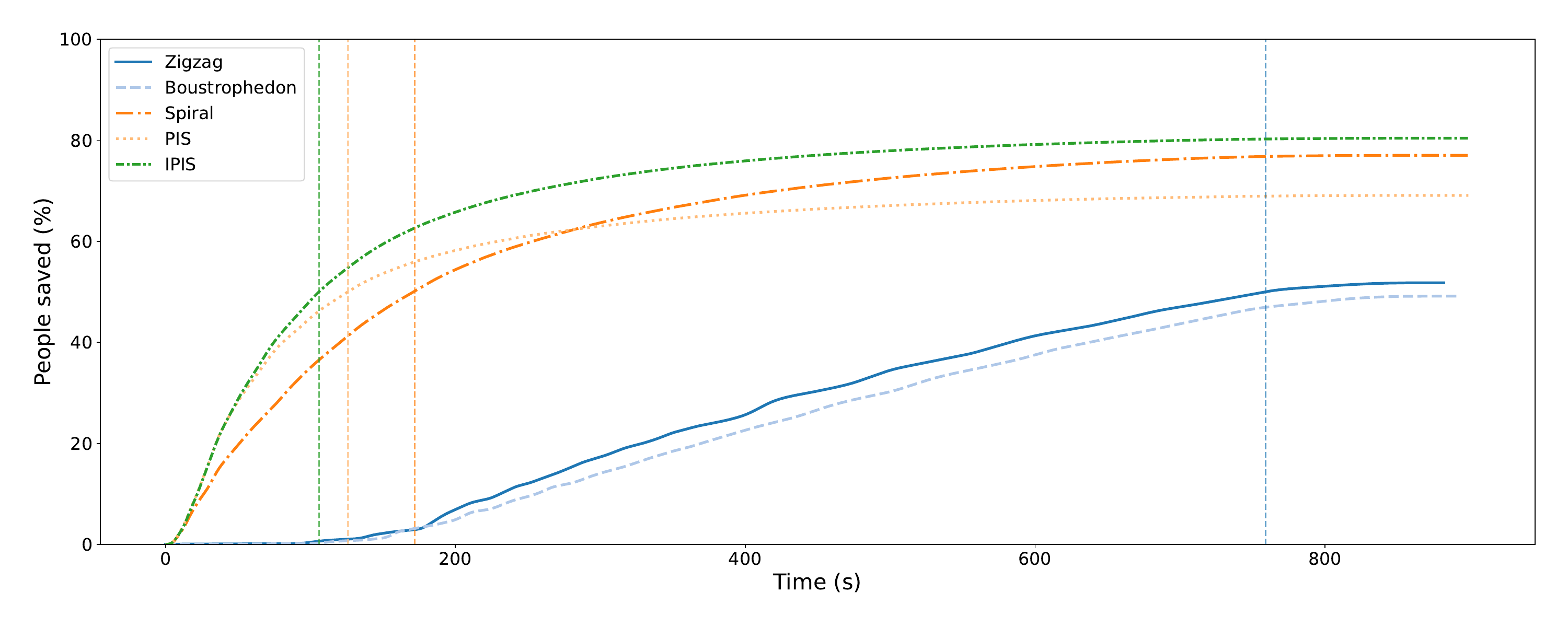}
    \label{fig:performance30}
  \end{subfigure}
  \caption{Comparison of performance of search methods for detection accuracies of $100\%$, $50\%$, and $30\%$. The perpendicular lines show the time that each method achieves saving 50\% of the people on average for all experiments.}
  \label{fig:performancombined}
\end{figure}
}

\subsection{Qualitative comparison of search methods}
\label{sec:qualitative-results}
In this section, we discuss the qualitative performance of the different search methods. The comparison of the methods is based on a single experiment that is representative of the conditions that can be encountered in the ocean. The goal of this comparison is to explain how the different methods work, showcase their strengths and shortcomings.

For the experiment, we chose the same UAV parameters as in section \ref{sec:quantitative-results}. The simulated UAV maintains an altitude of $20$ meters and an average moving speed of $20$ meter per second. The simulated weather conditions for the experiment are as follows: 

\begin{equation}
  \begin{aligned}
  V_w      &\sim \mathcal{N}(25, 5),   &\quad
  \theta_w &\sim \mathcal{N}(90, 5), \\
  V_c      &\sim \mathcal{N}(1, 0.2),  &\quad
  \theta_c &\sim \mathcal{N}(0, 5).
  \end{aligned}
\end{equation}
For the purpose of this comparison, we terminate the search based only on the battery life of the UAV, to evaluate how the methods work on a prolonged search horizon.

\begin{figure}[p]
  \centering

  \begin{subfigure}{0.48\linewidth}
    \centering
    \includegraphics[width=\linewidth]{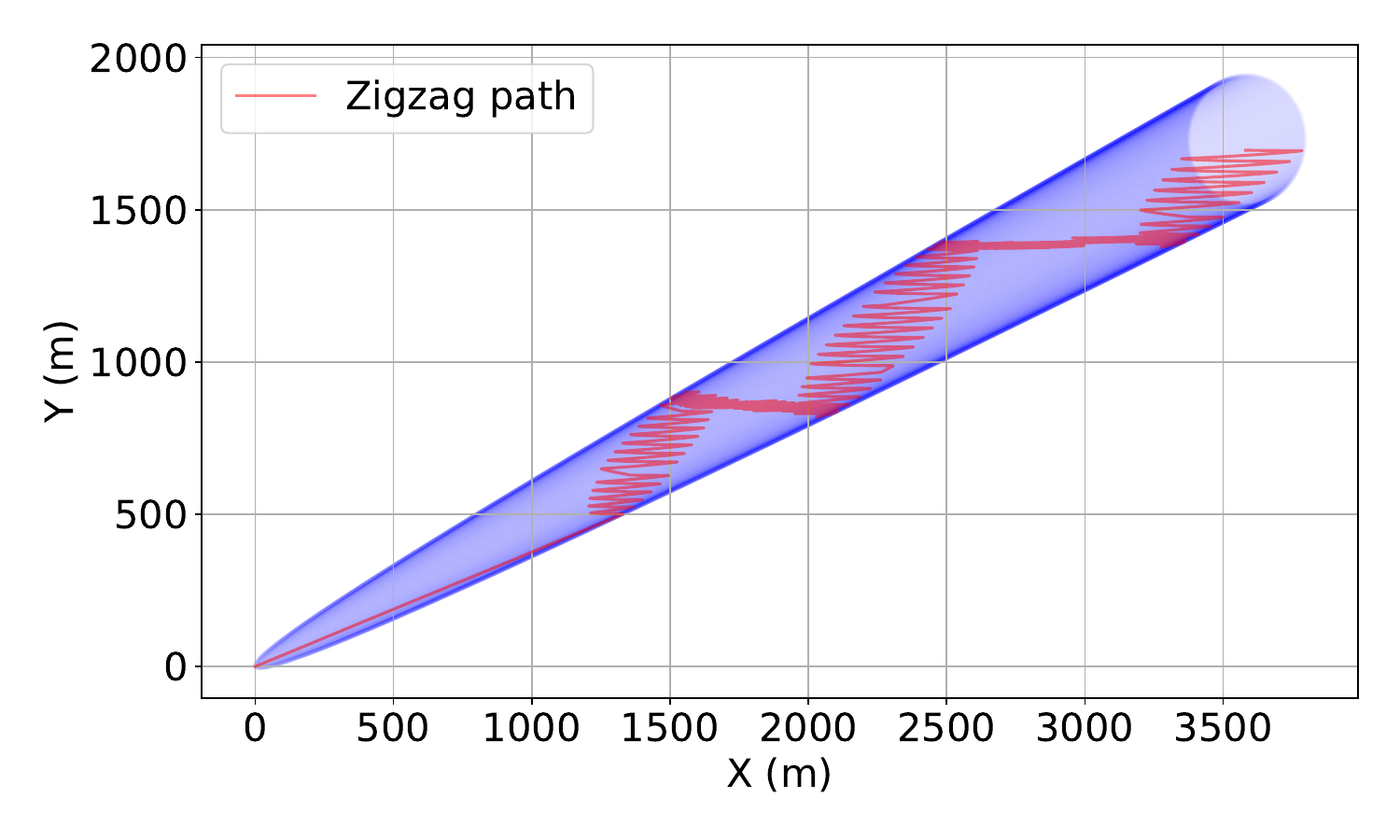}
    \caption{Zigzag method path.}
    \label{fig:zigzag-qualitative}
  \end{subfigure}\hfill
  \begin{subfigure}{0.48\linewidth}
    \centering
    \includegraphics[width=\linewidth]{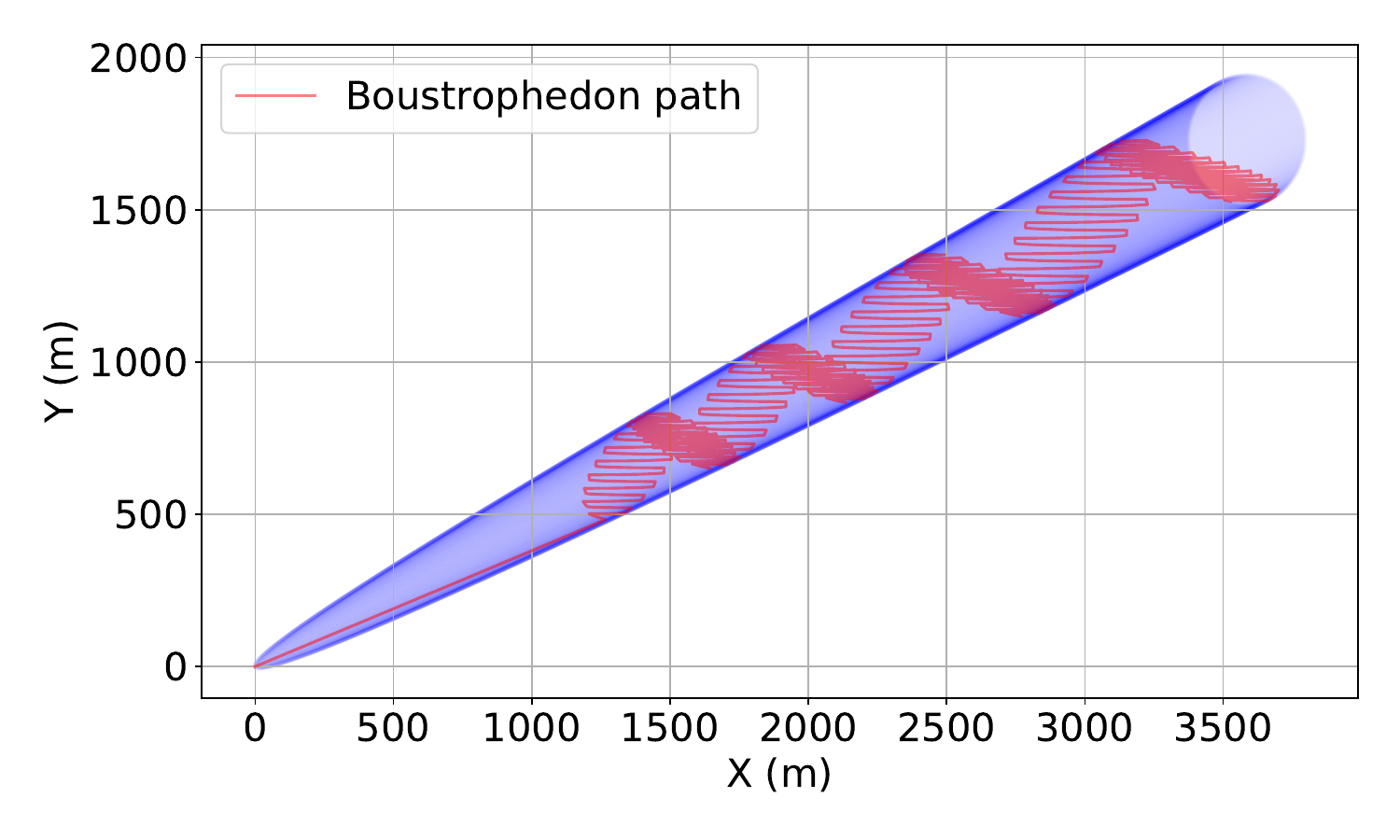}
    \caption{Boustrophedon method path.}
    \label{fig:boustrophedon-qualitative}
  \end{subfigure}

  \vspace{4mm}

  \begin{subfigure}{0.48\linewidth}
    \centering
    \includegraphics[width=\linewidth]{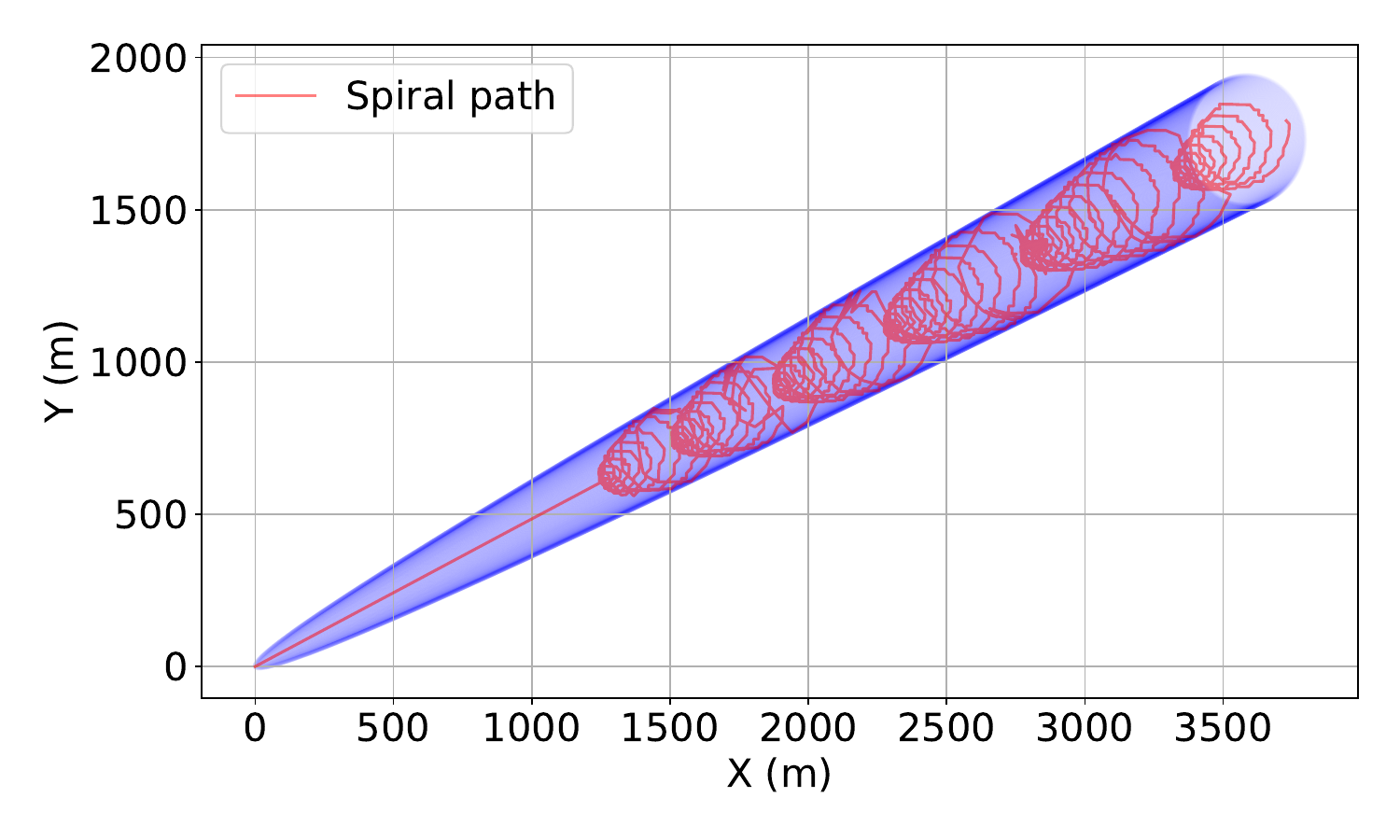}
    \caption{Spiral method path.}
    \label{fig:spiral-qualitative}
  \end{subfigure}\hfill
  \begin{subfigure}{0.48\linewidth}
    \centering
    \includegraphics[width=\linewidth]{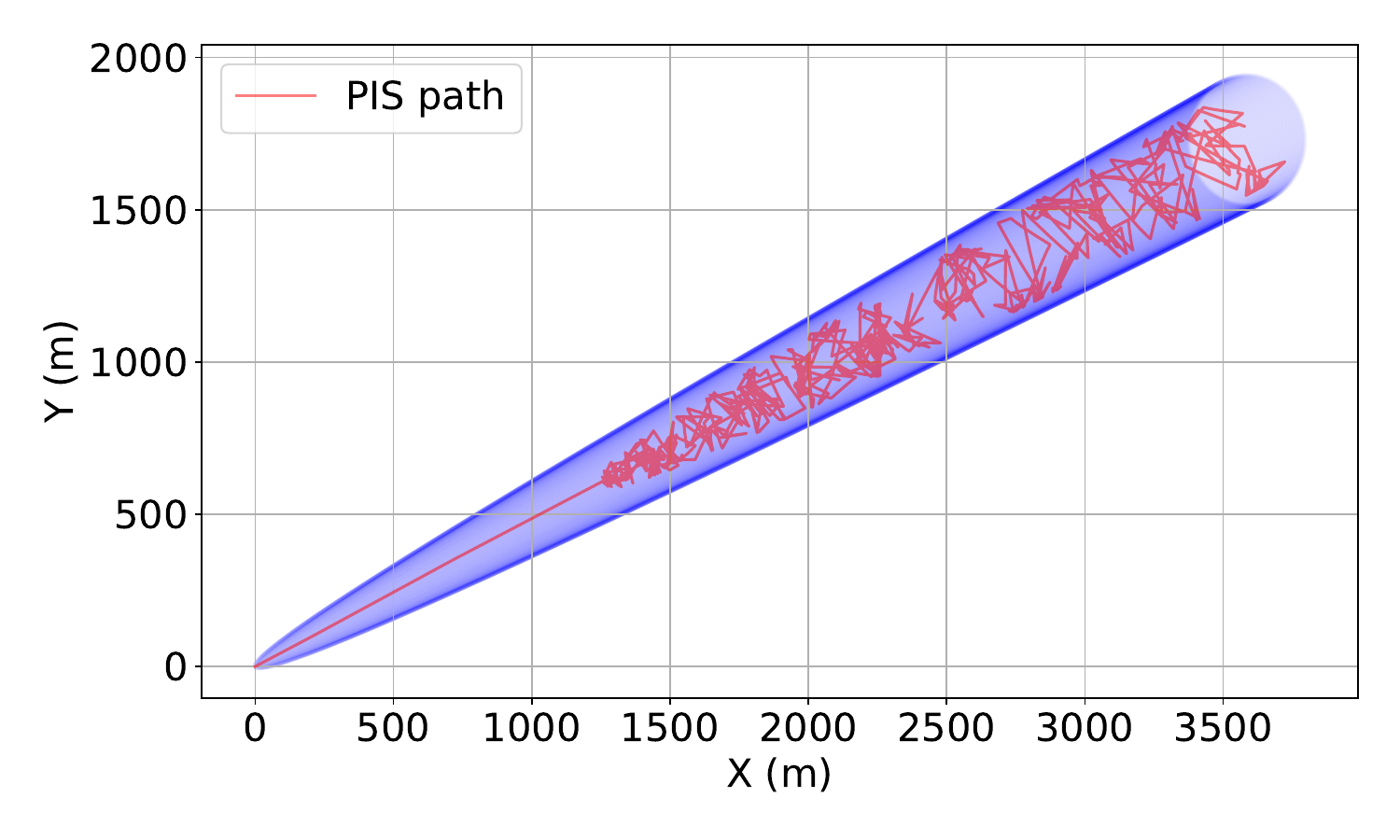}
    \caption{PIS method path.}
    \label{fig:bayes-qualitative}
  \end{subfigure}

  \vspace{4mm}

  \begin{subfigure}{0.48\linewidth}
    \centering
    \includegraphics[width=\linewidth]{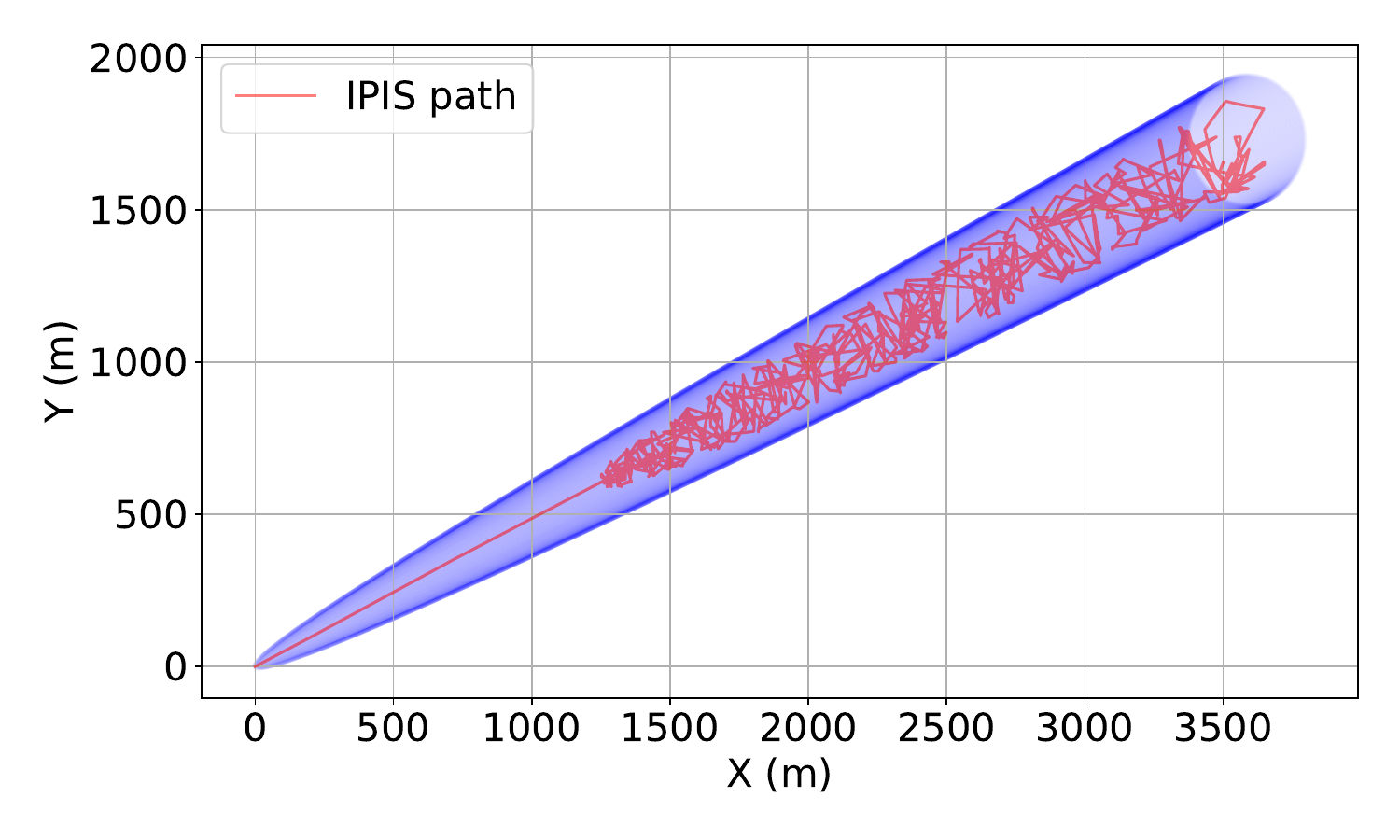}
    \caption{IPIS method path.}
    \label{fig:astar-qualitative}
  \end{subfigure}

  \caption{Qualitative performance of search methods for the same experiment parameters. The blue ellipses represent the predicted covariance ellipses from the EKF and the red lines are the path the UAV followed for each search method.}
  \label{fig:qualitative-comparison}
\end{figure}

In Figure \ref{fig:qualitative-comparison}, we present the results of the different search methods for the described experiment. Starting from the zigzag and boustrophedon methods, we can split their function in two steps. On the first step, both methods search throughout the area by opening the search radius. On the second step, the methods halt their forward movement and search laterally in a more localized area, as seen in Figures \ref{fig:zigzag-qualitative} and \ref{fig:boustrophedon-qualitative}. The interval of each step grows as the search continues, due to the increasing size of the search area. The boustrophedon method is faster than the zigzag method and is able to cover the whole area more times, explaining the increased performance we show in Table \ref{tab:results}.

In contrast to the zigzag and boustrophedon methods, the spiral method continuously covers the whole area by expanding the search from the center node of the ellipse to its edges. The search procedure has a single step that repeats, where the UAV moves in an outward spiraling motion until it reaches the edge of the search area, and then returns to the center to repeat the same pattern. This explains why the method performs better than the zigzag and boustrophedon methods in the {$F_{50}$} and MAUC metrics in Table \ref{tab:results}, since the probability of a person being near the center of the ellipse is higher than near the edges.

The probability informed methods follow paths that are not straightforward. We observe that both methods follow erratic paths during the search. On the one hand, the PIS method creates a more sparse path because it does not optimize it during the search. On the other hand, the IPIS method traverses faster through the nodes due to its optimized behavior, and it creates a more dense path. The methods are very reliable for particles that follow the predicted distribution but fail for particles that are edge cases.

\subsection{{IPIS performance}}

{
In this section we discuss the performance of the IPIS method to understand why it performs better than the other methods on average. The IPIS method utilizes the probability information that is produced by the EKF to identify the nodes with the highest probability of containing humans. It uses the optimal path between those nodes to efficiently traverse through the probability grid. As it traverses the area, it utilizes the Bayesian search theory to update the priors of each node based on the performance of its detection method. This way, it optimizes the search operation. 

The performance of the method is strongly influenced by particle dispersion rate. As shown in Section \ref{sec:quantitative-results}, the performance of the method degrades when the dispersion rate of the particles is high. This means that for extreme wind conditions, the method may not be performing optimally for the task at hand, but for lower wind conditions the success rate of finding a person can be higher than $97\%$.

The method is developed under the simplifying assumptions that the UAV's position is known exactly and that the camera systems never produce false negative predictions. However, these assumptions do not always hold in real-world scenarios. GPS and RTK-based localization systems heavily rely on sky visibility, rendering precise positioning difficult in adverse conditions, such as heavy rain. Similarly, the performance of detection systems varies with environmental factors, including illumination changes and out-of-distribution observations. The IPIS method prioritizes paths through higher probability nodes, but retains the ability to revisit an area multiple times, mitigating the uncertainty induced due to these factors. 
}

\section{Conclusion}
This paper presents a novel approach for detecting people that have fallen overboard vessels in open sea using a UAV. The search area estimator is able to accurately predict the area that a missing person might be in, by taking into account the weather measurements of the area from the vessel. A major advantage of the estimator is that it is computationally efficient, allowing it to run onboard a UAV, and eliminating the need for external computations and communication with the vessel or edge devices. Additionally, it can estimate the likelihood of a person being at a specific part of the area, enabling the development of more sophisticated path planning methods. Furthermore, the method is differentiable, allowing for future end-to-end learning-based search approaches.

The UAV then uses this prediction to search for the missing person. The estimator enables the UAV to employ the Zigzag, Boustrophedon, Spiral, PIS or IPIS methods to find the people with a probability of success $>80\%$, even if it initiates the search mission 20 minutes after the MOB incident. The selection of the search method can be done based on the other parameters of the UAV, such as the quality of the detection method. The IPIS method maintained a high success rate, regardless of the performance of the detection method that is used to locate the missing person.

{
  The search and rescue strategy that is proposed in this paper can decrease substantially the mortality rate of man-overboard incidents. The current strategies are dependent on manual search of people in water utilizing surface vehicles. The proposed strategy utilizes aerial vehicles that autonomously search the probable area of containing humans, effectively providing a more efficient and robust way of searching for them. Future steps would include large-scale field tests in collaboration with maritime organizations or coast guards.  
}

This paper paves the way for the development of more sophisticated MOB search methods. The IPIS method may be robust, but it consumes more computational power as the search area increases. Reinforcement learning techniques can be employed cooperatively with our prediction method, to create more efficient but equally robust search methods.


\begin{thebibliography}{00}
  \bibitem{stat4} CLIA, "Report on Operational Incidents 2009 to 2019" [Online]. Available: \url{https://cruising.org/resources/report-operational-incidents-2009-2019}. [Accessed: Apr. 29, 2025].

  \bibitem{stat0} Heggie, Travis W., and Tracey Burton-Heggie. "Death at sea: Passenger and crew mortality on cruise ships." International Journal of Travel Medicine and Global Health 8.4 (2020): 146-151.

  \bibitem{stat1} Statista, "Ocean shipping worldwide - statistics \& facts" [Online]. Available: \url{https://www.statista.com/topics/1728/ocean-shipping/}. [Accessed: Apr. 29, 2025].

  \bibitem{stat2} European Maritime Safety Agency. "Annual Overview of Marine Casualties and Incidents." EMSA, 2024, URL: \url{https://www.emsa.europa.eu/publications/item/5352-annual-overview-of-marine-casualties-and-incidents-2024.html}. [Accessed: Apr. 29, 2025].

  \bibitem{leeway1} Arthur A Allen, Robert Quincy Robe and ET Morton, "The Leeway of Persons-In-Water and Three Small Craft. Tech. rep", ANALYSIS and TECHNOLOGY INC NORTH STONINGTON CT, 1999.

  \bibitem{leeway2} Allen, Arthur A. "Leeway divergence." US Coast Guard Research and Development Center (2005).

  \bibitem{leeway3} Breivik, Øyvind, and Arthur A. Allen. "An operational search and rescue model for the Norwegian Sea and the North Sea." Journal of Marine Systems 69.1-2 (2008): 99-113.

  \bibitem{leewaysim1} Allen, Arthur, et al. "Field determination of the leeway of drifting objects." (2010).

  \bibitem{leewaysim2} Kako, Shin’ichiro, et al. "Inverse estimation of drifting-object outflows using actual observation data." Journal of oceanography 66 (2010): 291-297.

  \bibitem{leewaysim5} Brushett, Ben A., et al. "Application of leeway drift data to predict the drift of panga skiffs: Case study of maritime search and rescue in the tropical pacific." Applied ocean research 67 (2017): 109-124.

  \bibitem{leewaysim6} Richardson, Philip L. "Drifting in the wind: leeway error in shipdrift data." Deep Sea Research Part I: Oceanographic Research Papers 44.11 (1997): 1877-1903.

  \bibitem{leewaysim7} Zhang, Jinfen, et al. "Probabilistic modelling of the drifting trajectory of an object under the effect of wind and current for maritime search and rescue." Ocean Engineering 129 (2017): 253-264.

  \bibitem{leeway4} Breivik, Øyvind, et al. "Wind-induced drift of objects at sea: The leeway field method." Applied Ocean Research 33.2 (2011): 100-109.

  \bibitem{RWPT4} Ličer, Matjaž, et al. "Lagrangian modelling of a person lost at sea during the Adriatic scirocco storm of 29 October 2018." Natural Hazards and Earth System Sciences 20.8 (2020): 2335-2349.

  \bibitem{leeway7} Davidson, Fraser JM, et al. "Applications of GODAE ocean current forecasts to search and rescue and ship routing." Oceanography 22.3 (2009): 176-181.

  \bibitem{RL1} Wu, Jie, et al. "An autonomous coverage path planning algorithm for maritime search and rescue of persons-in-water based on deep reinforcement learning." Ocean engineering 291 (2024): 116403.

  \bibitem{opendrift} Dagestad, Knut-Frode, et al. "OpenDrift v1. 0: a generic framework for trajectory modelling." Geoscientific Model Development 11.4 (2018): 1405-1420.

  \bibitem{leeway8} Kim, Ji-Chang, et al. "Validation of opendrift-based drifter trajectory prediction technique for maritime search and rescue." Journal of Ocean Engineering and Technology 37.4 (2023): 145-157.

  \bibitem{RWPT1} Xiong, Weitao, P. H. A. J. M. Van Gelder, and Kewei Yang. "A decision support method for design and operationalization of search and rescue in maritime emergency." Ocean Engineering 207 (2020): 107399.

  \bibitem{RWPT2} Agbissoh Otote, Donatien, et al. "A decision-making algorithm for maritime search and rescue plan." Sustainability 11.7 (2019): 2084.

  \bibitem{RWPT3} Mou, Junmin, et al. "Cooperative MASS path planning for marine man overboard search." Ocean Engineering 235 (2021): 109376.

  \bibitem{sarops} Kratzke, Thomas M., Lawrence D. Stone, and John R. Frost. "Search and rescue optimal planning system." 2010 13th International Conference on Information Fusion. IEEE, 2010.

  \bibitem{gnome} Beegle-Krause, J. "General NOAA oil modeling environment (GNOME): a new spill trajectory model." International Oil Spill Conference. Vol. 2001. No. 2. American Petroleum Institute, 2001.

  \bibitem{medslik1} De Dominicis, Michela, et al. "MEDSLIK-II, a Lagrangian marine surface oil spill model for short-term forecasting–Part 1: Theory." Geoscientific Model Development 6.6 (2013): 1851-1869.

  \bibitem{medslik2} De Dominicis, M., et al. "MEDSLIK-II, a Lagrangian marine surface oil spill model for short-term forecasting–Part 2: Numerical simulations and validations." Geoscientific Model Development 6.6 (2013): 1871-1888.

  \bibitem{tracmass} Döös, Kristofer, Bror Jönsson, and Joakim Kjellsson. "Evaluation of oceanic and atmospheric trajectory schemes in the TRACMASS trajectory model v6. 0." Geoscientific Model Development 10.4 (2017): 1733-1749.

  \bibitem{parcels} Lange, Michael, and Erik van Sebille. "Parcels v0. 9: prototyping a Lagrangian ocean analysis framework for the petascale age." Geoscientific Model Development 10.11 (2017): 4175-4186.

  \bibitem{ariane1} Blanke, Bruno, and Stéphane Raynaud. "Kinematics of the Pacific equatorial undercurrent: An Eulerian and Lagrangian approach from GCM results." Journal of Physical Oceanography 27.6 (1997): 1038-1053.

  \bibitem{ariane2} Blanke, Bruno, et al. "Warm water paths in the equatorial Atlantic as diagnosed with a general circulation model." Journal of Physical Oceanography 29.11 (1999): 2753-2768.

  \bibitem{cms} Paris, Claire B., et al. "Connectivity Modeling System: A probabilistic modeling tool for the multi-scale tracking of biotic and abiotic variability in the ocean." Environmental Modelling \& Software 42 (2013): 47-54.

  \bibitem{search0} Li, Jiqiang, et al. "A survey of maritime unmanned search system: Theory, applications and future directions." Ocean Engineering 285 (2023): 115359.

  \bibitem{search1} Arkin, Esther M., Sándor P. Fekete, and Joseph SB Mitchell. "Approximation algorithms for lawn mowing and milling." Computational Geometry 17.1-2 (2000): 25-50.

  \bibitem{feraru} Feraru, Valeria Alexandra, Rasmus Eckholdt Andersen, and Evangelos Boukas. "Towards an autonomous UAS-based system to assist search and rescue operations in man-overboard incidents." 2020 IEEE international symposium on safety, security, and rescue robotics (SSRR). IEEE, 2020.

  \bibitem{dimos} Angelis, Dimosthenis, et al. "UAV Design for Fully Autonomous Man Overboard Detection." 2024 IEEE International Conference on Imaging Systems and Techniques (IST). IEEE, 2024.

  \bibitem{RL2} Zhan, Haowen, et al. "A reinforcement learning-based evolutionary algorithm for the unmanned aerial vehicles maritime search and rescue path planning problem considering multiple rescue centers." Memetic Computing 16.3 (2024): 373-386.

  \bibitem{RL3} Wu, Jie, et al. "A reinforcement learning-assisted search and rescue resource allocation decision-making approach for maritime emergencies." Computers \& Industrial Engineering 201 (2025): 110933.

  \bibitem{leeway6} Wu, Jie, Liang Cheng, and Sensen Chu. "Modeling the leeway drift characteristics of persons-in-water at a sea-area scale in the seas of China." Ocean engineering 270 (2023): 113444.

  \bibitem{montecarlo1} Melsom, Arne, et al. "Forecasting search areas using ensemble ocean circulation modeling." Ocean Dynamics 62 (2012): 1245-1257.

  \bibitem{leeway9} Meng, Sujing, et al. "A study on the leeway drift characteristic of a typical fishing vessel common in the Northern South China Sea." Applied Ocean Research 109 (2021): 102498.

  \bibitem{uav1} DJI. "DJI Air 3 Specifications." DJI, 2023. Available: https://www.dji.com/air-3/specs.

  \bibitem{uav2} DJI. "DJI Mavic 3 Classic Specifications." DJI, 2022. Available: https://www.dji.com/mavic-3-classic/specs.

  \bibitem{uav3} DJI. "DJI Matrice 30 Series Specifications." DJI Enterprise, 2022. Available: https://enterprise.dji.com/matrice-30/specs.

  \bibitem{uav4} DJI. "DJI Matrice 350 RTK Specifications." DJI Enterprise, 2023. Available: https://enterprise.dji.com/matrice-350-rtk/specs.

\end{thebibliography}
\end{document}